%% file: main.tex
\documentclass[11pt]{article}
\usepackage[final]{acl}
\usepackage{times}
\usepackage{latexsym}
\usepackage[T1]{fontenc}
\usepackage[utf8]{inputenc}
\usepackage{microtype}
\usepackage{inconsolata}
\usepackage{graphicx}
\usepackage{booktabs}
\usepackage{amsmath}
\usepackage{multirow}
\usepackage[table]{xcolor}
\usepackage{algorithm}
\usepackage{algpseudocode}
\usepackage{enumitem}
\usepackage{amssymb}
\usepackage{microtype}

\def \name{ReVisiT}
\newcommand{\ta}[1]{\textbf{#1}}
\newcommand{\tb}[1]{\underline{#1}}
\newcommand{\tc}[1]{\text{#1}}
\newcommand{\ca}[1]{\cellcolor{gray!20}\textbf{#1}}
\newcommand{\cb}[1]{\cellcolor{gray!20}\underline{#1}}
\newcommand{\cc}[1]{\cellcolor{gray!20}\text{#1}}

\title{Revisit What You See: Revealing Visual Semantics\\in Vision Tokens to Guide LVLM Decoding}

\author{
  Beomsik Cho \quad
  Jaehyung Kim \\
  Yonsei University \\
  \texttt{\{bscho333,jaehyungk\}@yonsei.ac.kr}
}

\begin{document}
\maketitle

\input{Tex/Abstract}
\input{Tex/Introduction}
\input{Tex/Related}
\input{Tex/Motivation}
\input{Tex/Method}
\input{Tex/Experiment}
\input{Tex/Conclusion}
\newpage
\input{Tex/Limitation_Ethical}

\bibliography{custom}
\input{Tex/Appendix}

\end{document}

%% file: Tex/Abstract.tex
\begin{abstract}
Large Vision Language Models (LVLMs) achieve strong performance across multimodal tasks by integrating visual perception with language understanding. 
However, how vision information contributes to the model’s decoding process remains under-explored, as reflected in frequent hallucinations.
Through a series of analyses, we found that (i) vision tokens provide meaningful visual information even when hallucinations occur, and (ii) their semantics are encoded in the textual space and become explicit under appropriate vocabulary constraints.
Building on these observations, we propose \textbf{\name{}}, a simple training-free decoding method that guides text generation in LVLMs by  \textbf{Re}ferencing \textbf{Visi}on \textbf{T}okens.
Our approach leverages the semantic information embedded within vision tokens by projecting them into the text token distribution.
Specifically, \name{} dynamically selects the most relevant vision token at each decoding step via context-aware constrained divergence minimization.
Then, \name{} uses its constrained projection to refine the output distribution to better incorporate visual semantics. 
Across five benchmarks on recent LVLMs, \name{} achieves competitive or superior results to state-of-the-art decoding baselines while reducing computational cost by up to $2\times$.\footnote{Code: \url{https://github.com/bscho333/ReVisiT}.}
\end{abstract}

%% file: Tex/Introduction.tex
\section{Introduction}
\label{intro}

\input{Figures/models}

With the recent success of Large Language Models (LLMs) \citep{touvron2023llama,achiam2023gpt,team2023gemini}, Large Vision-Language Models (LVLMs) have emerged as powerful multimodal architectures that integrate visual perception with language understanding and text generation \citep{dai2023instructblipgeneralpurposevisionlanguagemodels,liu2024improved,zhu2023minigpt,bai2025qwen2,zhu2025internvl3}.
Typically, LVLMs encode visual inputs into the LLM decoder’s text embedding space as \textit{vision tokens}, which are then processed alongside text tokens during decoding.
This way of treating vision tokens as static auxiliary contexts, similar to retrieval-augmented generation \citep{lewis2020retrieval, gao2023retrieval} in LLMs, enables the construction of complex multimodal systems in a relatively simple manner.
However, this approach is often insufficient to capture the unique characteristic and role of vision tokens as sole carriers of visual information, as denoted by hallucinations frequently observed in LVLM’s text outputs \citep{leng2024vcd,favero2024m3id,huo2024sid,rohrbach2018chair,li2023pope,guan2024hallusionbench}.
To mitigate this, research has begun to expand the understanding of vision tokens along with their additional utilization \citep{jiang2025interpreting, jiang2025devils}.
Yet, it is still under-explored \textit{how vision tokens influence the decoding process of LVLMs} and \textit{what kinds of textual semantics they encode}.

To investigate this direction, we conduct a series of analyses on vision tokens, leading to two key insights.
First, by examining the output distributions of LVLMs particularly at hallucinated steps, we find that ground-truth objects often remain among the model’s high-probability candidates and this visual grounding signal in the output distribution comes from vision tokens. 
Second, we observe that the textual semantics of vision tokens is initially obscured, but the interpretable and well-aligned semantics could be discovered with the proper constraints to focus on specific subset. 
Here, we interpret the semantics of vision tokens by projecting them into the text token distribution using the LLM decoder’s language modeling head, as both are embedded in the same space. 
Overall, these findings suggest that vision tokens intrinsically encode object-level semantics, though such signals typically remain latent under conventional unconstrained decoding.

Building on this observation, we propose \textbf{\name{}}, a simple yet effective decoding strategy that
guides text generation in LVLMs by \textbf{Re}ferencing \textbf{Visi}on \textbf{T}okens.
At each step, \name{} selects the most relevant vision token based on \textit{context-aware constrained divergence} from the current output distribution.
Then, \name{} uses its constrained projection as reference to refine token-level logits.
This activates the intrinsic semantics of vision tokens without training, architectural modifications, external modules, or multi-pass inference.

As highlighted in Fig.~\ref{fig:models}, \name{} consistently improves performance across six model sizes spanning three architectures, demonstrating scalability and broad applicability across LVLM families.
To be specific, on three hallucination benchmarks: HallusionBench~\citep{guan2024hallusionbench}, CHAIR~\citep{rohrbach2018chair}, and POPE~\citep{li2023pope}, \name{} reduces hallucinations robustly on three representative model architectures: LLaVA-1.5~\citep{liu2024improved}, Qwen2.5-VL~\citep{bai2025qwen2}, and InternVL3-8B~\citep{zhu2025internvl3}.
Beyond hallucination, we further validate its effectiveness on general-purpose benchmarks, including VQAv2~\citep{vqa_Antol_2015_ICCV} and MMMU~\citep{vqa_yue2024mmmu}, indicating that the semantics surfaced by our context-aware constrained projection generalize across tasks and architectures.

\noindent In summary, our contributions are as follows:
\begin{itemize}[leftmargin=4mm]
    \vspace{-0.02in}
    \item[$\circ$]
    We provide quantitative and qualitative evidence that vision tokens intrinsically encode interpretable semantics exposed when projected over \emph{semantically coherent vocabulary subset}.
    \vspace{-0.05in}
    \item[$\circ$]
    We introduce \textbf{\name{}}, a training-free LVLM decoding method that references vision tokens via \emph{context-aware constrained divergence}.
    \vspace{-0.05in}
    \item[$\circ$]
    We validate across \emph{six model sizes (7B to 32B) spanning three architectures}, reducing hallucinations and yielding gains on five benchmarks, while maintaining inference cost.
\end{itemize}

%% file: Figures/models.tex
\begin{figure}[t]
\centering
\includegraphics[width=1.0\linewidth]{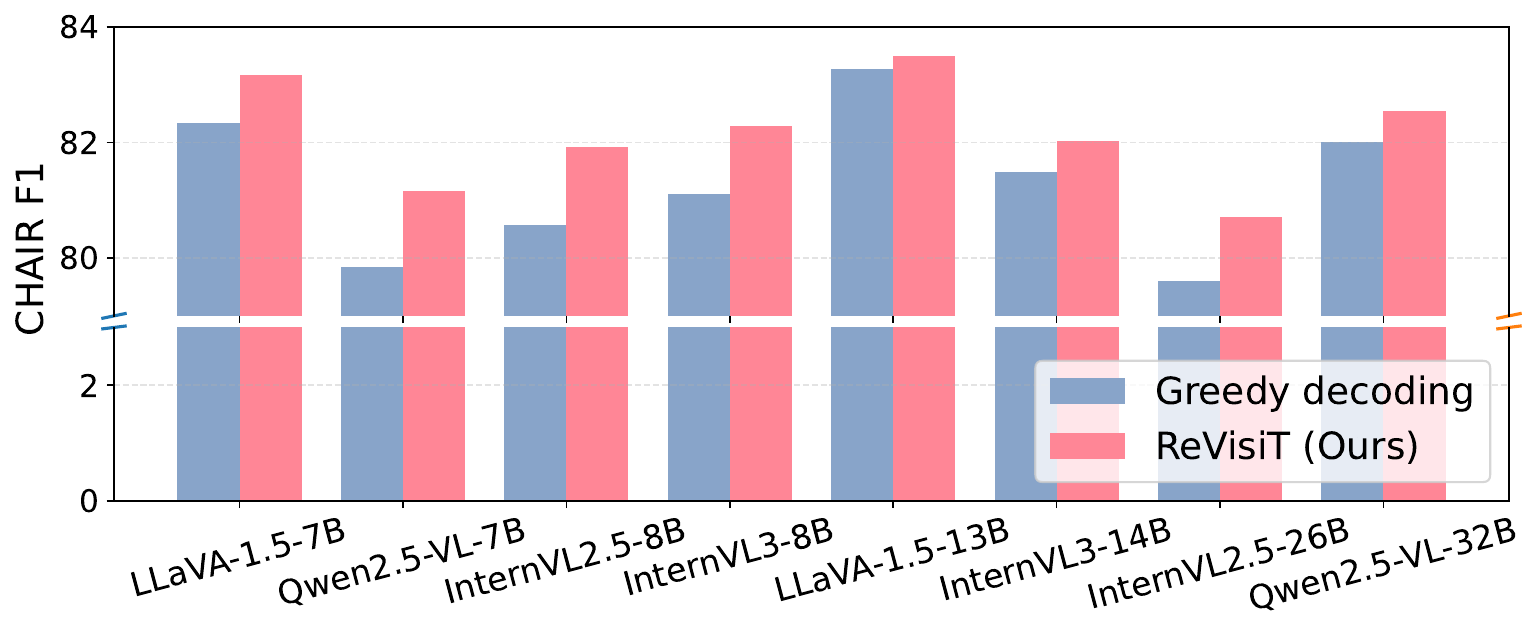}
\caption{
    \textbf{\name{} across various model sizes and architectures.} 
    We evaluate on the CHAIR benchmark and report object-level F1 computed over CHAIR object matches.
    Consistent improvements hold across different size buckets (7--8B / 13--14B / 26--32B) and architectures, demonstrating scalability and broad applicability.
}
\vspace{-0.2in}
\label{fig:models}
\end{figure}

%% file: Tex/Related.tex
\section{Related Works}
\label{sec:related}

\paragraph{Aligning visual inputs with language generation.}
Prior attempts to align visual inputs with language generation in LVLMs can be broadly grouped into \emph{pre-alignment}, \emph{intra-alignment}, and \emph{post-alignment}, depending on whether the alignment signal is injected before, during, or after decoding.
Pre-alignment methods modify model architectures and training pipelines so that visual and textual modalities are better aligned before inference, for example by refining vision encoders and cross-modal projection modules~\citep{Qwen-VL,chen2024internvl,lu2024ovis,you2023ferret}.
Post-alignment approaches operate on the generated sequence and correct hallucinations or factual errors using auxiliary revisors or verification pipelines conditioned on the image and caption~\citep{zhou2023lure,yin2024woodpecker}.
Intra-alignment methods intervene during decoding, for example by contrastively comparing logits under perturbed inputs to suppress language-prior-driven hallucinations~\citep{leng2024vcd,favero2024m3id,kim2024code}, by recalibrating attention to visual tokens using vision-aware adjustments derived from attention patterns~\citep{gong2024damro,woo2024avisc,huo2024sid,kang2025see}, or by introducing auxiliary visual-sensitive distributions or penalties and modifying decoding rules to better ground tokens in the visual input~\citep{chen2024halc,huang2024opera,wang2024deco}.
Across these approaches, visual information typically influences decoding implicitly through hidden states, logits, or attention scores.

\input{Figures/motivation_overall}

\paragraph{Interpreting vision representation via projection into language space.}
Complementary to alignment frameworks, another line of work focuses on interpreting internal representation by projecting them into the vocabulary space using logit-lens-style probes~\citep{logitlens,chuang2023dola}.
Recently, this approach has been extended to vision representations in LVLMs to analyze vision-token semantics and support feature editing~\citep{jiang2025interpreting,jiang2025devils}.
However, prior work typically collapses them into scalar confidence or importance scores for diagnosing or reweighting internal model components.
In contrast, we treat the projected distribution itself as a \emph{context-aware constrained} signal that directly modulates next-token prediction.
This positions our method at the intersection of projection-based interpretability and decoding-time visual grounding: we explicitly expose the textual semantics of individual vision tokens in the language space and use them as an interpretable guidance signal for generation.

%% file: Figures/motivation_overall.tex
\begin{figure*}[h]
\centering
\includegraphics[width=0.9\textwidth]{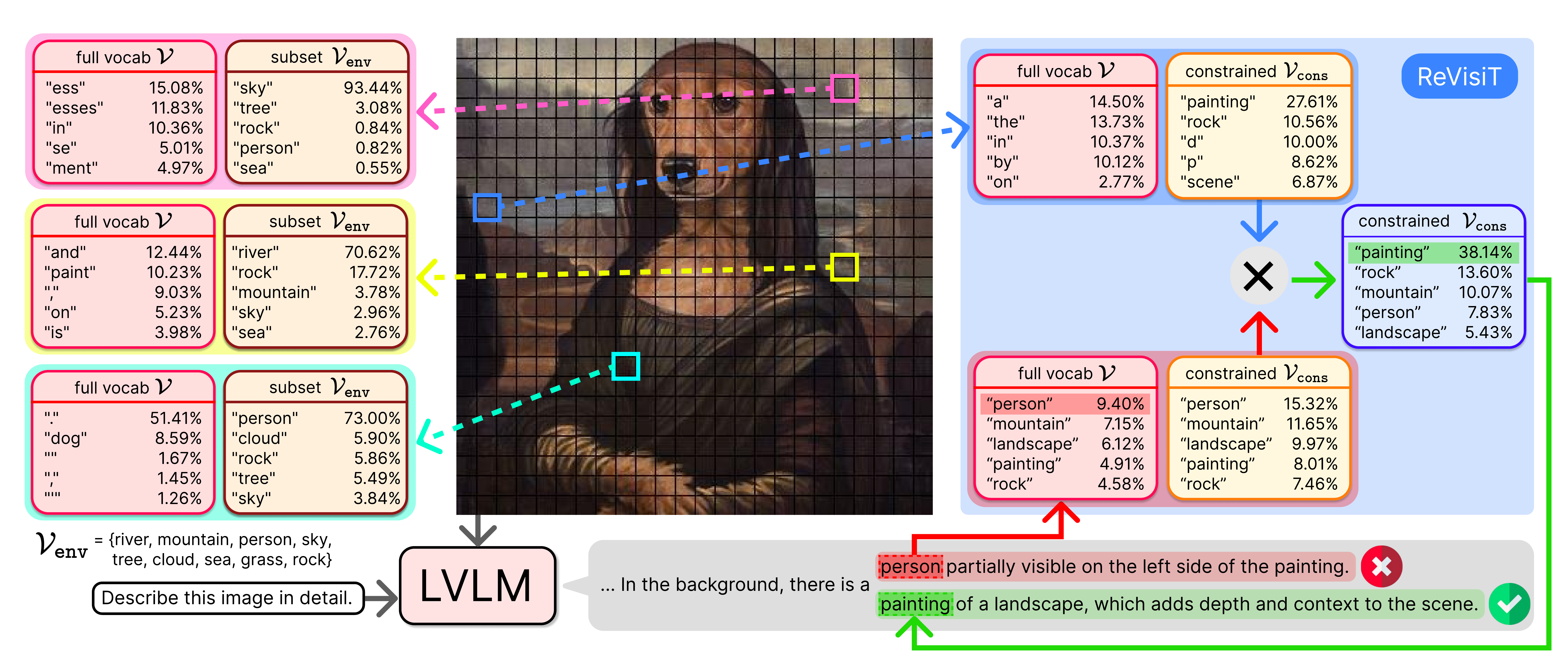}
\caption{
    \textbf{Motivation.} We qualitatively analyzed various vision tokens with LLaVA-1.5-7B \citep{liu2024improved}. Dotted arrows represent vision token projection over specified vocabulary set. For each box, representing text token distribution, we annotated top-5 probable text tokens. Left part illustrates the effectiveness of vocabulary constraint, whereas right part shows the distribution shift during \name{}. See Appendix~\ref{appendix:motivation-fig} for a detailed discussion.
}
\vspace{-0.1in}
\label{fig:motivation}
\end{figure*}

%% file: Tex/Motivation.tex
\section{Empirical Analysis of How Vision Tokens Influence LVLM Decoding}\label{sec:motivation}
In this section, we investigate whether vision tokens encode interpretable semantics in LVLM's text token space that can directly inform its decoding process.
To conduct systematic analyses, we adopt the CHAIR benchmark \citep{rohrbach2018chair}, as it provides a principled definition of object existence based on 80 MS-COCO \citep{lin2014mscoco} object categories with 403 synonyms in total; 
for each sample, the set of synonyms of annotated object categories is regarded as ground-truth (GT) objects, while the remaining category synonyms are regarded as hallucinated (Hal) objects.  
We refer to a \textit{hallucinated step} as a decoding step where generated token corresponds to an object in this MS-COCO synonym set but is not part of the ground-truth annotations for the given image.

Building on this definition, we sample 500 images from the MS-COCO validation set and generate captions with \textit{Qwen2.5-VL-7B}~\citep{bai2025qwen2} using vanilla greedy decoding with a maximum of 512 tokens and the standard CHAIR prompt, \texttt{``Please describe this image in detail''}.

\subsection{Evidence of visual grounding in output distributions with vision tokens}
\label{sec:mot2.1}
To examine how vision tokens affect LVLM decoding, 
we first ask whether ground-truth (GT) objects remain among the model’s high-probability predictions at hallucinated steps. 
We quantify this using \emph{at-least-one recall@$k$}, which measures whether at least one GT object synonym appears among the top-$k$ predictions of a given distribution (formal definition in Appendix~\ref{appendix:metric-def}). 
Among the 500 generated captions, 176 captions contain at least one hallucinated step, yielding 190 hallucinated steps for step-level analysis.
For each hallucinated step, we report \emph{at-least-one recall@$k$} on the model’s output distribution at that step, excluding the greedy (top-1) prediction from the candidate set.

\input{Figures/motivation_exp1}

The results are summarized in Fig.~\ref{fig:motivation_exp1}. 
Here, GT objects frequently remain accessible in output distribution: in 63.2\% of hallucinated steps, at least one GT object appears within top-50 candidates, and this recall rises to 95.8\% within top-500. 
Given the vocabulary size of 151{,}665 tokens, top-500 corresponds to only 0.33\% of the full vocabulary.\footnote{Step-selection details and 95\% confidence intervals are provided in Appendix~\ref{appendix:exp1}.}
A qualitative example is also presented in Fig.~\ref{fig:motivation} (right), where the model incorrectly selects ``person'' as the greedy prediction, yet visually grounded alternatives such as ``mountain,'' ``landscape,'' and ``painting'' appear among top-5 candidates.
This indicates that relevant visual candidates are assigned high probabilities in the output distribution, showing that LVLM indeed possesses the correct visual information even when hallucinations occur.

\input{Tables/motivation_exp2}

We further conduct a fine-grained comparison between the occurrences of GT and hallucinated object subsets during the decoding. 
Specifically, for each generated token, we track the maximum probability among objects in each subset. 
Then, we report the average of these per-step maxima as \textit{GT Max Avg} for GT object subsets and \textit{Hal Max Avg} for hallucinated object subsets, respectively (formal definition in Appendix~\ref{appendix:exp2}).
To investigate the contribution of vision tokens by isolating them, we additionally report \textit{w/o image} setting where we remove all vision tokens and perform teacher forcing with the previously generated caption as input while keeping decoding otherwise identical. 

As shown in Table~\ref{tab:motivation_exp2} (top), with vision tokens present, GT objects receive higher probability mass than hallucinated ones on average (GT $>$ Hal); removing the vision tokens reverses the ordering (Hal $>$ GT), confirming a visual grounding contribution from them.
Interestingly, even for hallucinated samples, the similar results are observed (Table~\ref{tab:motivation_exp2} (middle)) which indicates that grounding signal still exists in general. 
However, when focusing on the specific hallucinated steps, the reversed ordering (Hal $>$ GT) persists even with the presentation of image, indicating that \textit{language priors} (\textit{i.e.}, textual co-occurrence biases from pre-training \citep{leng2024vcd, favero2024m3id}) can dominate the prediction at specific positions and it leads to the hallucination in LVLM decoding.\footnote{Wilcoxon signed-rank test results are provided in Appendix~\ref{appendix:exp2}.}

In summary, these results indicate that, \textit{even when hallucinations occur, LVLMs consistently retain strong visual grounding signals in their output distributions}, but not selected during decoding.

\subsection{Object-level semantics in vision tokens}
\label{sec:mot2.2}

To identify the source of visual grounding signals, we focus on \textit{vision tokens}, the sole carriers of visual information in LVLMs.
Vision tokens share the same embedding space with text tokens and are autoregressively processed during decoding like them.
This motivates probing their representations by projecting them through the model’s language-modeling head to obtain the text token distribution over the vocabulary, analogous to logit-lens and early-exit analyses for text token embeddings \citep{logitlens,chuang2023dola,wang2024deco}. 
Specifically, for each of 500 MS-COCO validation images, we project each vision token's final-layer hidden state to obtain the text token distribution, yielding 173,173 vision tokens and their projected distributions (details in Appendix~\ref{appendix:exp3}).

\input{Figures/motivation_exp3}

Then, to assess the presence of visual grounding signals in vision tokens, we compute token-level \emph{at-least-one recall@$k$} for the projected distribution, similar to analyses in Fig.~\ref{fig:motivation_exp1}.
The quantitative results are summarized in Fig.~\ref{fig:motivation_exp3}.
Quantitatively, across 173,173 vision tokens, \emph{at-least-one recall@$k$} is low under the full vocabulary: only \textbf{2.03\%} of projections surface a GT synonym at top-1, increasing to \textbf{21.72\%} at top-30.
Qualitative inspection also indicates that tokens corresponding to background regions often yield generic function words, with object labels rarely appearing (red boxes on the left side of Fig.~\ref{fig:motivation}).
These observations indicate that naive vision-token projections with the full vocabulary are diffuse and difficult to interpret.

In contrast, constraining the projection to \textit{semantically coherent} vocabulary subsets reveals substantially clearer alignment with visual semantics. 
When restricting the prediction space to the CHAIR-derived object subset (176 single-token synonyms covering 62 categories), \emph{at-least-one recall@$k$} on the \emph{same} 173{,}173 projections increases sharply to \textbf{40.44\%} at top-1—about $20\times$ higher than the naive full-vocabulary projection (2.03\%). At top-30, recall reaches \textbf{89.24\%} while inspecting only $30/176\!\approx\!17\%$ of the subset.\footnote{Token-level \emph{at-least-one recall@$k$} statistics, including 95\% confidence intervals, are reported in Appendix~\ref{appendix:exp3}.}
For comparability, both settings evaluate GT synonyms from the same 62 categories. 
Moreover, when we manually define domain-specific subsets, such as environment-related terms (\textit{e.g.}, ``river,'' ``mountain,'' and ``tree''), the projected distribution aligns with visually salient regions and reveals object labels that remain hidden under the full-vocabulary setting (brown boxes on the left side of Fig.~\ref{fig:motivation}). 

These findings show that \textit{vision tokens encode rich object-level semantics and properly designed semantic constraints can reliably reveal these latent visual signals}. 

For clarity, the subsets used so far are for controlled analysis and qualitative illustration, whereas the actual \name{} uses the adaptive context-aware candidate set introduced next in Eq.~\eqref{eq:our_apc}.

%% file: Figures/motivation_exp1.tex
\begin{figure}[t]
{
    \centering
        \includegraphics[width=1.0\linewidth]{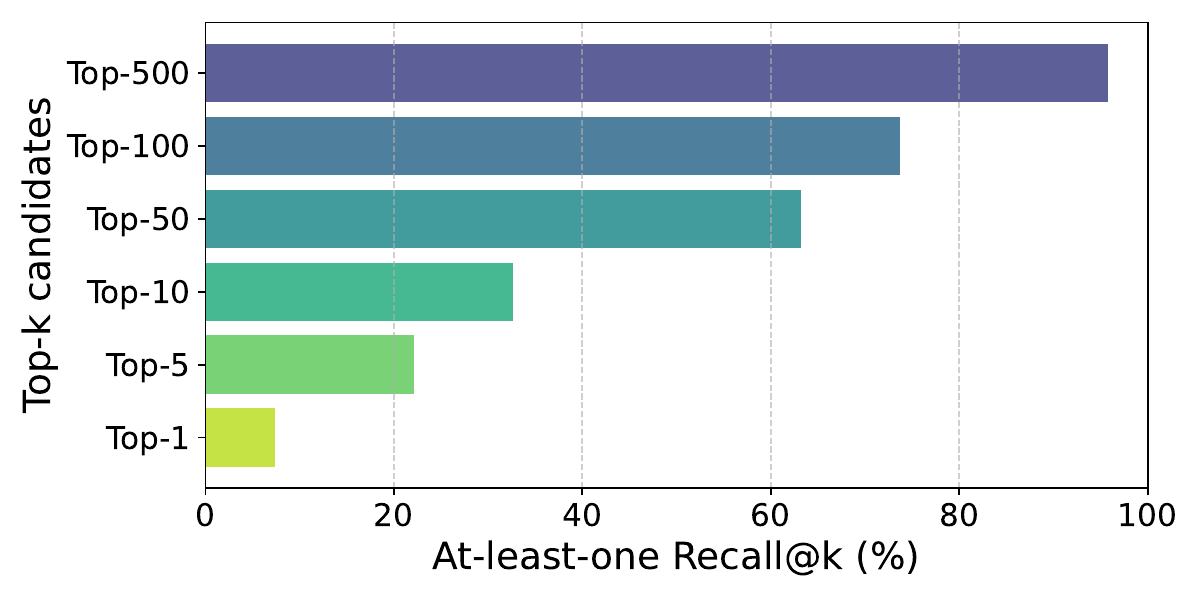}
        \vspace{-0.3in}
        \caption{\textbf{Ground-truth objects frequently remain in top-probability predictions.}
        At 190 hallucinated steps, at least one GT object is recalled in 95.8\% of cases within top-500 predictions, corresponding to only 0.33\% of full vocabulary (151,665 tokens).}
    \label{fig:motivation_exp1}
    \vspace{-0.1in}
}
\end{figure}

%% file: Tables/motivation_exp2.tex
\begin{table}[t]
    \small
    \centering
    \caption{\textbf{Comparison of Hal Max Avg and GT Max Avg.} 
        The results for (i) all output tokens across all 500 CHAIR samples (88,706 tokens), 
        (ii) all output tokens within hallucinated samples (176 samples, 35,116 tokens), 
        and (iii) hallucinated decoding steps within hallucinated samples (176 samples, 190 steps). 
        The higher value between Hal and GT is highlighted in \textbf{bold}.}
    \vspace{-0.05in}
    \label{tab:motivation_exp2}
    \setlength{\tabcolsep}{4pt}
    \resizebox{1.0\linewidth}{!}{
    \begin{tabular}{lc|cc}
        \hline
        {Analysis Setting} & {Condition} & {GT Max Avg} & {Hal Max Avg} \\
        \hline
        \multirow{2}{*}{{All tokens}} 
            & w/ image & \textbf{24.49} & 20.24 \\
            & w/o image & 21.83 & \textbf{22.66} \\
        \hline
        {All tokens} 
            & w/ image & \textbf{24.42} & 21.74 \\
            {(in hallucinated samples)} & w/o image & 21.73 & \textbf{23.88} \\
        \hline
        \multirow{2}{*}{{Hallucinated tokens}} 
            & w/ image & 32.00 & \textbf{36.44} \\
            & w/o image & 30.26 & \textbf{35.23} \\
        \hline
\end{tabular}}
\vspace{-0.1in}
\end{table}

%% file: Figures/motivation_exp3.tex
\begin{figure}[t]
{
    \centering
        \includegraphics[width=0.9\linewidth]{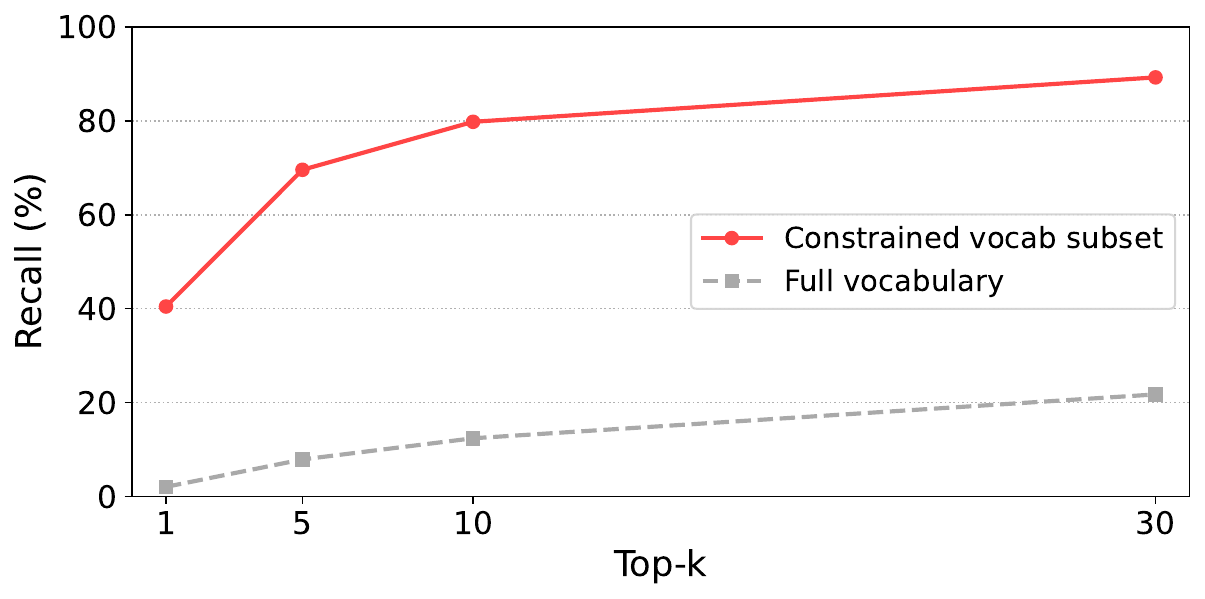}
        \vspace{-0.15in}
        \caption{\textbf{Vision token projection.}
        Vision tokens reveal rich semantics when projected over semantically coherent subsets.}
    \label{fig:motivation_exp3}
    \vspace{-0.1in}
}
\end{figure}

%% file: Tex/Method.tex
\section{\name{}: Referencing Vision Tokens to Guide Text Generation with LVLMs}
\label{sec:method}

Based on insights from Sec.~\ref{sec:motivation}, we propose \name{}, a decoding method that leverages these inherent visual semantics to enhance visual grounding during text generation.
\name{} uses the vanilla output distribution as a context-aware constraint to surface the relevant visual semantics encoded in vision tokens, and integrates the revealed information to refine the next-token distribution. 
Specifically, \name{} consists of three main steps:  
(1) adaptively constraining the vocabulary subset based on the vanilla output distribution,  
(2) projecting the hidden states of vision tokens onto this subset and selecting the most relevant one, and  
(3) refining the output distribution through log-space fusion/addition and normalization. 
The algorithm and illustration of \name{} are presented in Appendix~\ref{appendix:revisit}.

\textbf{Preliminary.} Let $\mathbf{x}$ and $\mathbf{v}$ denote the input text tokens (e.g., query) and vision tokens, respectively. 
Specifically, the vision tokens $\mathbf{v}$ are generated from the given input image by forwarding it through the vision encoder and cross-modality projector sequentially~\citep{liu2024improved,bai2025qwen2}.
Then, our goal is to generate output text tokens $\mathbf{y}$ using LVLM $\mathcal{M}$; following the auto-regressive nature of LLMs, output tokens $\mathbf{y}$ (\textit{e.g.,} response) are sequentially generated, \textit{i.e.}, $y_t \sim \mathcal{M}(\cdot|\mathbf{v},\mathbf{x},\mathbf{y}_{<t})$.
Let us assume that LVLM $\mathcal{M}$ has $L$ decoding layers and text token vocabulary $\mathcal{V}$, and denote its hidden state of $l$-th layer as $h^{l}$.
The output distribution is parameterized by model weights $\theta$ and denoted as $p_{\theta}$.
Then, at time step $t$, output probability over $\mathcal{V}$ is calculated as:
\begin{equation}
    p_{\theta}(h_{T+t-1}^L,\mathcal{V}) = \mathrm{softmax}_\mathcal{V} \left( \phi(h_{T+t-1}^L) \right),
    \label{eq:softmax}
\end{equation}
where $T = |\mathbf{v}| + |\mathbf{x}|$, $\phi$ is the language modeling head that maps the final hidden state of LVLM into logits over vocabulary $\mathcal{V}$, and $\mathrm{softmax}_{\mathcal{V}}$ denotes a softmax operation over $\mathcal{V}$ such that $\sum_{w\in\mathcal{V}} p_{\theta}(w|h_{T+t-1}^L,\mathcal{V}) =1$.
Then, the output token $y_t$ is sampled from the obtained distribution: 
\begin{equation}
    y_t \sim p_{\theta}(h_{T+t-1}^L, \mathcal{V}).\label{eq:output}
\end{equation}

\input{Figures/overview}

\subsection{Context-aware vocabulary subset construction}
At each decoding timestep $t$, the LVLM $\mathcal{M}$ generates an output distribution $p_{\theta}(h_{T+t-1}^L, \mathcal{V})$ over the entire vocabulary space $\mathcal{V}$.
However, as observed in Fig.~\ref{fig:motivation} and discussed in Sec.~\ref{sec:mot2.1}, this distribution is often diffusely spread across many irrelevant tokens.
Therefore, we dynamically constrain a vocabulary $\mathcal{V}$ into subset $\mathcal{V}_{\texttt{cons}}^t$ more suitable to capture the plausible semantic space of the output distribution.
By restricting the output space to a smaller, contextually focused subset, this constraint reduces semantic distraction and facilitates more precise integration of visual information.
To be specific, motivated by the Adaptive Plausibility Constraint (APC)~\citep{li2022contrastive}, 
we adaptively define the vocabulary subset $\mathcal{V}_{\texttt{cons}}^t$ as below:
\begin{equation}
\resizebox{0.9\columnwidth}{!}{$
    \mathcal{V}_{\texttt{cons}}^t = 
    \left\{ 
    w \in \mathcal{V} : \;
    \begin{aligned}
        &p_{\theta}(w|h_{T+t-1}^L, \mathcal{V}) \\
        &\geq \alpha \cdot \max_{w'} p_{\theta}(w'|h_{T+t-1}^L, \mathcal{V})
    \end{aligned}    
    \right\},
    \label{eq:our_apc}
$}
\end{equation}
where $\alpha \in (0, 1)$ controls the tightness of the selected candidate set. Smaller $\alpha$ yields a looser subset, while larger $\alpha$ yields a tighter subset.
This subset $\mathcal{V}_{\texttt{cons}}^t$ serves as a dynamically adapted output space, 
over which all subsequent vision token projections and refinements are restricted, thereby enhancing visual grounding without additional computation.

\subsection{Vision token selection with projection}
Given $\mathcal{V}_{\texttt{cons}}^t$ at each decoding timestep $t$, we next describe how vision token hidden states $h_i^j$ are projected and selected within this constrained space.
Let $\mathcal{J}$ denote the set of candidate decoder layers where vision token hidden states are selected.  
Specifically, for each $h_i^j$, where $i$ denotes the index of the vision tokens and $j$ denotes the candidate layer, we obtain the projected distribution over $\mathcal{V}_{\texttt{cons}}^t$:
\begin{equation}
\resizebox{0.9\columnwidth}{!}{$
    p_{\theta}(h_i^j, \mathcal{V}_{\texttt{cons}}^t),
    \quad \text{for} \quad i \in \{0, \ldots, |\mathbf{v}|-1\}, \; j \in \mathcal{J}.\label{eq:vision_project}
$}
\end{equation}
Intuitively, this distribution represents the likelihood of each candidate token within $\mathcal{V}_{\texttt{cons}}^t$ conditioned on the vision token embedding $h_i^j$. 
Next, to identify the vision token most relevant to the current decoding context, we compare each vision token distribution $p_{\theta}(h_i^j, \mathcal{V}_{\texttt{cons}}^t)$ against the vanilla output distribution $p_{\theta}(h_{T+t-1}^L, \mathcal{V}_{\texttt{cons}}^t)$ by computing the Jensen-Shannon Divergence (JSD):
\begin{equation}
\resizebox{\columnwidth}{!}{$
    (i^*, j^*) = \arg \underset{i, j}\min  \; \mathrm{JSD}\left( p_{\theta}(h_{T+t-1}^L, \mathcal{V}_{\texttt{cons}}^t) \parallel p_{\theta}(h_i^j, \mathcal{V}_{\texttt{cons}}^t) \right).
$}
\end{equation}
Here, $i^*$ and $j^*$ denote the indices of the selected vision token and decoder layer, respectively. 
We adopt JSD as the divergence metric due to its symmetric properties and bounded output range. 

\textbf{Remark.} For the efficiency, the vision token projections over the full vocabulary, $p_{\theta}(h_i^j, \mathcal{V})$, are precomputed once before decoding begins.  
Then, at each timestep $t$, we apply slicing and masking to adapt the cached projections to the dynamically constrained subset $\mathcal{V}_{\texttt{cons}}^t$, enabling efficient selection without additional forward computation.

\subsection{Logit refinement with projected token distribution of selected vision token}
\label{sec:method_logit_refinement}
After identifying the most contextually relevant vision token at each decoding step,  
we then refine the vanilla output distribution guided by the reference visual grounding signal from the selected vision token.
This refinement admits a KL-regularized, Product-of-Experts (PoE)-style interpretation; we provide the derivation in Appendix~\ref{appendix:revisit_theory}.
Specifically, given the vanilla output distribution on $\mathcal{V}_{\texttt{cons}}^t$, $p_{\theta}(y_t \mid h_{T+t-1}^L, \mathcal{V}_{\texttt{cons}}^t)$, and the selected vision-token distribution on the same constrained set, $p_{\theta}(y_t \mid h_{i^*}^{j^*}, \mathcal{V}_{\texttt{cons}}^t)$, we combine them via log-space addition:
\begin{equation}
\resizebox{0.8\columnwidth}{!}{$
    \label{eq:log-space-addition}
    l_{\texttt{ReVisiT}}(y_t)
    =
    \begin{aligned}
        & \log p_\theta(y_t \mid h^L_{T+t-1}, \mathcal{V}^t_{\texttt{cons}}) \\
        & + \log p_\theta(y_t \mid h^{j^*}_{i^*}, \mathcal{V}^t_{\texttt{cons}}).
    \end{aligned}
$}
\end{equation}
The final distribution is then defined as
\begin{equation}
\resizebox{0.95\columnwidth}{!}{$
    p_{\texttt{fin}}(y_t) =
    \begin{cases}
        \mathrm{softmax}_{\mathcal{V}^t_{\texttt{cons}}}\!\big(l_{\texttt{ReVisiT}}(y_t)\big), & y_t \in \mathcal{V}^t_{\texttt{cons}},\\
        0, & \text{otherwise}.\label{eq:final_fusion}
    \end{cases}
$}
\end{equation}
The next output token $y_t$ is sampled from the resulting distribution $p_{\texttt{fin}}$.
This refinement mechanism allows the decoding process to dynamically integrate visual grounding signals extracted from vision tokens, without introducing additional inference passes or architectural modifications.

%% file: Figures/overview.tex
\begin{figure*}[h]
\centering
\includegraphics[width=1.0\textwidth]{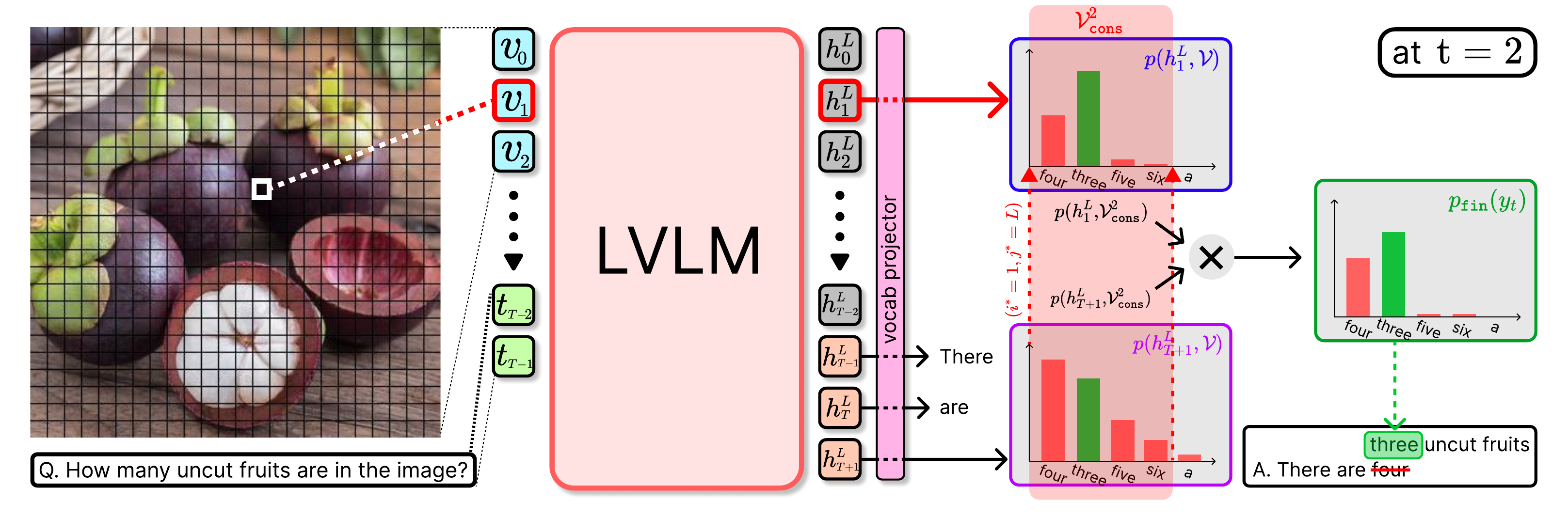}
\vspace{-0.2in}
\caption{
    \textbf{An overview of \name{}.} 
    At each decoding step, \name{}
    (1) constrains the vocabulary $\mathcal{V}$ to $\mathcal{V}_{\texttt{cons}}^t$,
    (2) projects vision token embeddings into $\mathcal{V}_{\texttt{cons}}^t$ and selects most relevant token, and
    (3) refines the final output distribution.
    \name{} leverages vision tokens to serve as reference signals for decoding, enhancing visual grounding without additional forward passes or external supervision.
}
\vspace{-0.1in}
\label{fig:overview}
\end{figure*}

%% file: Tex/Experiment.tex
\section{Experiments}
\label{sec:experiment}

\input{Tables/HallusionBench}
\input{Tables/CHAIR}

\subsection{Setups}
\paragraph{Models and baselines.} 
We mainly evaluate three representative state-of-the-art LVLMs: \textit{LLaVA-1.5-7B} \citep{liu2024improved} as a widely adopted conventional baseline, \textit{Qwen2.5-VL-7B} \citep{bai2025qwen2}, and \textit{InternVL3-8B} \citep{zhu2025internvl3}.
As baselines, we adopt various intra-alignment decoding strategies to compare the effectiveness of our method in enhancing visual grounding.
(a) \textit{Greedy decoding} is a standard decoding baseline, and
(b) \textit{DoLa} \citep{chuang2023dola}
improves factuality by contrasting output logits from earlier and later decoder layers at each timestep, and has been widely adopted for hallucination mitigation in LVLMs despite being originally proposed for LLMs.
For intra-alignment baselines,
(c) \textit{VCD} \citep{leng2024vcd} mitigates hallucination by perturbing visual inputs with Gaussian noise and enforcing consistency between the original and perturbed outputs;
(d) \textit{M3ID} \citep{favero2024m3id} contrasts outputs with and without visual input to alleviate language prior bias and enhance visual grounding; 
(e) \textit{CODE}~\citep{kim2024code} leverages the model's self-generated image description as a contrasting reference to alleviate hallucination; 
(f) \textit{SID} \citep{huo2024sid} applies contrastive decoding on low-attention vision patches to refine visual grounding.

\paragraph{Implementation details.}
In entire experiments, we apply deterministic \textbf{greedy decoding} (\textit{i.e.}, $\text{temperature}=0$) with a maximum output length of 512 tokens for \textit{all decoding methods}.
Regarding baselines, we follow the default hyperparameters provided by their original implementations.
For \name{}, we set the hyperparameters as follows: for LLaVA-1.5-7B and InternVL3-8B, we select vision tokens mostly from the last decoder layer (``last''), while for Qwen2.5-VL-7B, we select vision tokens mostly from all even-numbered decoder layers (``all'').
The threshold $\alpha$ is chosen per task type. 
More details are in Appendix~\ref{appendix:implementation-details}.

\input{Tables/POPE_summary}
\input{Tables/VQA}

\subsection{Main results}
We evaluate our method on five benchmarks that jointly span three axes:
(i) \emph{generative} vs.\ \emph{discriminative} settings,
(ii) \emph{manually designed} vs.\ \emph{LLM-aided} evaluation, and
(iii) \emph{hallucination-focused} vs.\ \emph{general VQA} tasks,
allowing us to assess the robustness and generalizability of the model's descriptive and decision-making abilities.
Detailed benchmark and metric descriptions are in Appendix~\ref{appendix:dataset-details}, and full quantitative results are in Appendix~\ref{appendix:quanti}.

\textbf{HallusionBench}~\citep{guan2024hallusionbench}
probes image–context reasoning and entangled hallucinations with an LLM-based judge.
In the visual-dependent setting (Table~\ref{tab:hallusionbench}), \name{} attains the best score on four out of five metrics for each LVLM; on LLaVA-1.5-7B and InternVL3-8B it is second-best on the remaining metric, while on Qwen2.5-VL-7B its \textit{EaAcc} stays within 1.5 points of the best method.
In particular, \name{} improves \textit{qAcc} over greedy decoding by approximately \(75\%\) on LLaVA-1.5-7B and \(28\%\) on InternVL3-8B, indicating that referencing vision tokens yields more reliable visual grounding even when image and context provide conflicting cues.

\textbf{CHAIR}~\citep{rohrbach2018chair}
measures hallucination rates in image captioning by comparing generated captions to ground-truth object annotations from the MS-COCO~\citep{lin2014mscoco} dataset.
We report sentence-level (\textit{CHAIR$_S$}) and instance-level (\textit{CHAIR$_I$}) hallucination scores on 500 sampled images, and additionally include an object-level precision/recall \textit{F1} computed over CHAIR object matches to reflect both hallucination suppression and grounded object coverage.
As shown in Table~\ref{tab:CHAIR}, \name{} achieves the best performance on InternVL3-8B and consistently ranks within top-2 across all three metrics on LLaVA-1.5-7B and Qwen2.5-VL-7B.
This indicates that \name{} reduces hallucinations without collapsing into overly conservative captions, achieving a more balanced trade-off between hallucination suppression and object coverage, as reflected by improvements in all three metrics.

\textbf{POPE}~\citep{li2023pope}
evaluates object presence with binary questions across three datasets (MS-COCO, A-OKVQA~\citep{schwenk2022okvqa}, and GQA~\citep{hudson2019gqa}) and three query types (random, popular, adversarial).
We report averaged \textit{Accuracy}, \textit{Precision}, and \textit{F1} over all datasets and query types.
As summarized in Table~\ref{tab:POPE_summary}, \name{} achieves the best \textit{Accuracy} and \textit{F1} on both LLaVA-1.5-7B and InternVL3-8B, and ranks second on both metrics for Qwen2.5-VL-7B, yielding top-2 performance in every model–metric pair and indicating robust gains in discriminative grounding across architectures.

\textbf{VQAv2}~\citep{goyal2017vqav2}
measures VQA correctness under inter-annotator variability via consensus accuracy over 10 annotations.
As shown in Table~\ref{tab:mmmu_vqav2} (left), \name{} improves over greedy decoding on both LLaVA-1.5-7B and Qwen2.5-VL-7B, achieving the highest accuracy among all decoding methods on these models.
In contrast to baselines that exhibit substantial performance drops and ranking changes, \name{} maintains consistently competitive performance across all three models, indicating the robustness of \name{} across architectures without causing VQA failures.

\textbf{MMMU}~\citep{vqa_yue2024mmmu}
is a college-level benchmark spanning 30 disciplines of multimodal question answering, requiring substantial subject knowledge and reasoning in both multiple-choice and open-ended formats.
As shown in Table~\ref{tab:mmmu_vqav2} (right), \name{} achieves the highest accuracy for all three LVLMs, demonstrating its ability to generalize across diverse LVLMs and task setups.

\textbf{Overall.} 
Across these five benchmarks, \name{} remains consistently competitive across diverse evaluation settings, from object-level grounding to knowledge-intensive multimodal reasoning.
This consistency suggests that the proposed context-aware constrained divergence captures a broadly useful decoding signal that improves visual grounding while reducing hallucinations, rather than a benchmark-specific heuristic.
Such versatility, achieved without any additional training or multi-pass inference, underscores the efficiency and generalizability of leveraging intrinsic vision-token semantics for robust multimodal alignment.

\subsection{Additional analyses}
\label{sec:experiment_additional}
\paragraph{Inference speed improvement.} 
To evaluate the inference efficiency of \name{} compared to baseline decoding strategies, we measure token-level inference latency (ms/token) for LLaVA-1.5-7B, Qwen2.5-VL-7B, and InternVL3-8B.
\begin{figure}[t]
    \includegraphics[width=\linewidth]{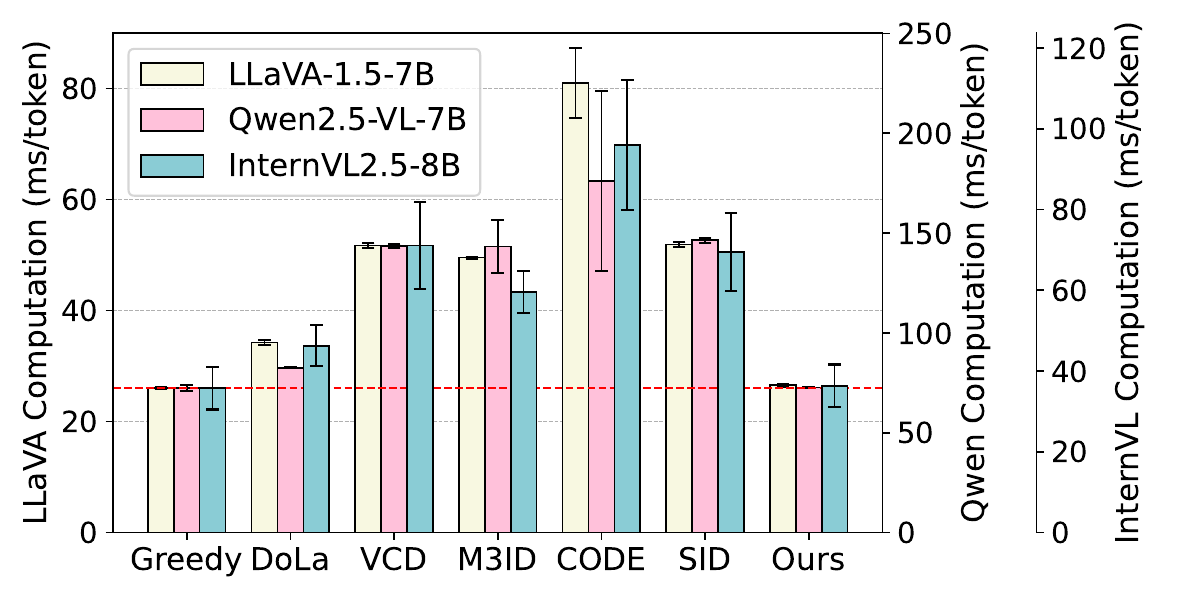}
    \vspace{-0.25in}
    \caption{\textbf{Inference speed.}
    Comparison of per-token inference latency across different decoding strategies for LLaVA-1.5-7B (left y-axis), Qwen2.5-VL-7B (right y-axis), and InternVL3-8B (outer right y-axis), with standard deviations visualized as error bars.
    }
    \vspace{-0.1in}
    \label{fig:estimation}
\end{figure}

As shown in Figure~\ref{fig:estimation}, \name{} remains comparable to vanilla greedy decoding, increasing runtime by only \(0.6\%\) to \(2.0\%\) across three LVLMs.
In contrast, methods that require additional forward passes (e.g., VCD, M3ID, SID, CODE) \emph{approach or exceed} \(2\times\) the greedy runtime, while even DoLa, which only adds a projection on intermediate layers at each step, incurs \(14\%\) to \(30\%\) overhead across all three models.
These results highlight a key advantage of \name{}: it enhances visual grounding with negligible computational overhead, unlike prior methods that significantly compromise inference efficiency.
See Appendix~\ref{appendix:computation} for measurement setup, hardware configuration, and full numerical results.

\begin{figure}
    \includegraphics[width=\linewidth]{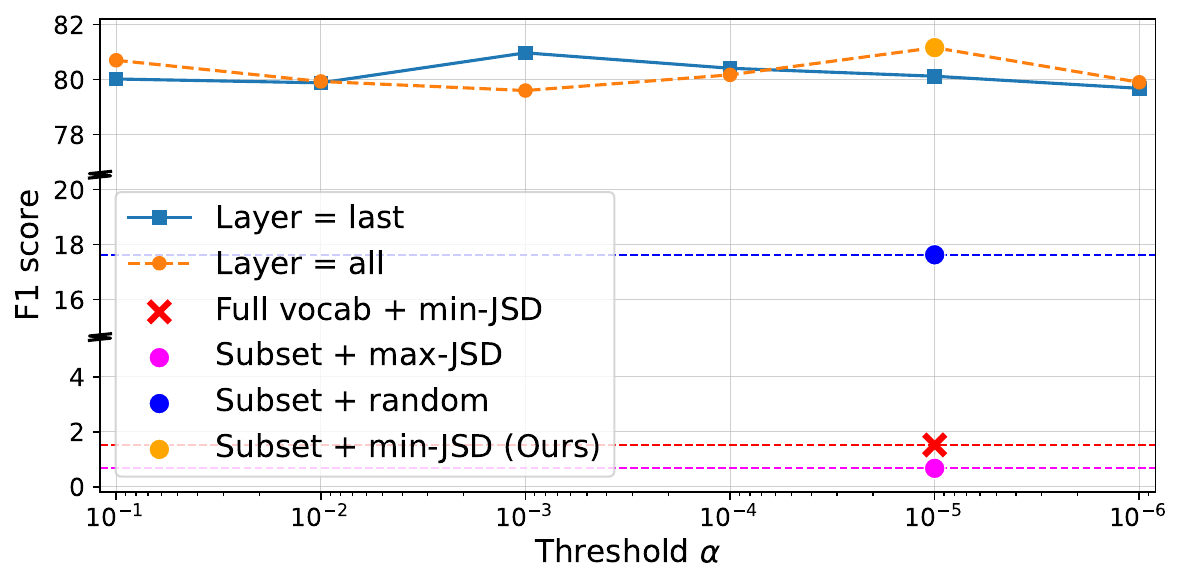}
    \vspace{-0.23in}
    \caption{\textbf{Ablation study.}
    We evaluate the impact of vocabulary subset constraint, vision token selection, layer scope, and threshold $\alpha$ variation on CHAIR benchmark.
    }
    \label{fig:ablation}
    \vspace{-0.1in}
\end{figure}

\paragraph{Ablation study.}
We ablate three factors on Qwen2.5-VL under the CHAIR benchmark (Figure~\ref{fig:ablation}): 
(1) vision-token selection criterion (min-/random/max-JSD), 
(2) vocabulary constraint (subset vs.\ full), and
(3) layer scope (all layers vs.\ last layer) with 
threshold $\alpha$.
 
\textit{(1) Selection:}
Under a constrained (subset) vocabulary, max-JSD collapses (F1 $=0.67$) and random underperforms (F1 $=17.63$), while our min-JSD attains the best performance (F1 $=81.16$), showing the effectiveness of a context-aware constrained divergence for token selection.

\textit{(2) Vocabulary constraint:}
Even with the same min-JSD selection, projecting over the full vocabulary fails (F1 $=1.52$), exhibiting a degenerate behavior where vision-token projections are referenced off-target. 
This confirms that constraining the candidate set to context-relevant tokens is crucial; qualitative failures are provided in Appendix~\ref{appendix:wo-vocab}.

\textit{(3) Layer scope \& threshold:}
Across $\alpha$, the all-layers variant peaks at $\alpha=10^{-5}$ (F1 $=81.16$), while the last-layer variant peaks at $\alpha=10^{-3}$ (F1 $=80.97$).
Overall sensitivity to $\alpha$ is modest (within $\approx\!1.5$ F1 of each peak), indicating that \emph{all-layers + vocabulary subset + min-JSD} is a robust setting on CHAIR.
Additionally, on VQAv2, accuracy increases from 65.93 at $\alpha=10^{-1}$ to 67.60 at $\alpha=10^{-2}$, but decreases to 64.00 at $\alpha=10^{-4}$, again suggesting that overly loose candidate sets are detrimental.

Comprehensive ablation results, including the full tables and additional design-choice experiments, are reported in Appendix~\ref{appendix:ablation}.

%% file: Tables/HallusionBench.tex
\begin{table*}[h]
\begin{center}
\begin{small}
\setlength{\tabcolsep}{3pt}
\caption{
    \textbf{HallusionBench}.
    Accuracies (\%) on five metrics: \textit{qAcc} (Question-pair accuracy), \textit{fAcc} (Figure accuracy), \textit{aAcc} (All accuracy), \textit{EaAcc} (Easy question accuracy), and \textit{HaAcc} (Hard question accuracy).
    Higher scores ($\uparrow$) indicate better performance.
    The best results in each metric are \textbf{bolded}, and the second-best are \underline{underlined}.
}
\vspace{-0.1in}
\label{tab:hallusionbench}
\resizebox{1.0\textwidth}{!}{
\begin{tabular}{lccccccccccccccc}
    \toprule
      \multirow{2}{*}[-0.5ex]{\textbf{Method}}
        & \multicolumn{5}{c}{\textbf{LLaVA-1.5-7B}}
        & \multicolumn{5}{c}{\textbf{Qwen2.5-VL-7B}}
        & \multicolumn{5}{c}{\textbf{InternVL3-8B}} \\
     \cmidrule(lr){2-6}\cmidrule(lr){7-11}\cmidrule(lr){12-16}
         & qAcc $\uparrow$ & fAcc $\uparrow$ & aAcc $\uparrow$ & EaAcc $\uparrow$ & HaAcc $\uparrow$
         & qAcc $\uparrow$ & fAcc $\uparrow$ & aAcc $\uparrow$ & EaAcc $\uparrow$ & HaAcc $\uparrow$
         & qAcc $\uparrow$ & fAcc $\uparrow$ & aAcc $\uparrow$ & EaAcc $\uparrow$ & HaAcc $\uparrow$ \\
     \midrule
    Greedy & \tc{11.55} & \tc{22.61} & \tc{48.22} & \tc{44.40} & \tc{48.54}
           & \tc{29.96} & \tc{34.78} & \tc{61.08} & \ta{65.70} & \tc{53.65}
           & \tc{25.99} & \tc{33.91} & \tc{56.35} & \tc{61.73} & \tc{48.54} \\
    DoLa   & \tc{13.36} & \tc{22.61} & \tb{49.24} & \ta{45.85} & \tb{49.27}
           & \tc{27.08} & \tc{30.87} & \tc{58.04} & \tc{62.09} & \tc{51.46}
           & \tb{32.49} & \tb{37.39} & \ta{59.90} & \tb{65.70} & \tb{51.82} \\
    VCD    & \tb{14.44} & \tc{22.61} & \tc{47.55} & \tc{44.77} & \tc{47.45}
           & \tc{29.60} & \tb{35.22} & \tc{61.25} & \tb{64.62} & \tb{54.74}
           & \tc{28.88} & \tc{35.65} & \tc{57.36} & \tc{62.82} & \tc{49.64} \\
    M3ID   & \tc{11.91} & \tb{23.04} & \tc{48.22} & \tc{43.68} & \tb{49.27}
           & \tc{30.32} & \tc{33.48} & \tc{60.07} & \tc{62.82} & \tc{54.38}
           & \tc{28.16} & \tc{33.91} & \tc{57.02} & \tc{60.65} & \tc{50.73} \\
    CODE   & \tc{10.47} & \tc{20.43} & \tc{46.36} & \tc{41.52} & \tc{47.81}
           & \tc{28.88} & \tc{31.74} & \tc{57.53} & \tc{57.76} & \tc{54.74}
           & \tc{28.52} & \tc{36.52} & \tc{56.68} & \tc{62.09} & \tc{49.27} \\
    SID    & \tc{12.27} & \tc{21.30} & \tc{48.22} & \tc{44.77} & \tc{48.18}
           & \tb{32.13} & \ta{36.09} & \tb{61.59} & \ta{65.70} & \tc{55.11}
           & \tc{31.05} & \tc{36.09} & \tc{58.04} & \tc{61.73} & \ta{52.19} \\
    \ca{Ours}
           & \ca{20.22} & \ca{26.52} & \ca{51.44} & \cb{45.13} & \ca{54.38}
           & \ca{32.85} & \ca{36.09} & \ca{61.76} & \cc{64.26} & \ca{56.93}
           & \ca{33.21} & \ca{39.13} & \cb{59.39} & \ca{66.07} & \ca{52.19} \\
    \bottomrule
\end{tabular}
}
\end{small}
\end{center}
\vspace{-10pt}
\end{table*}

%% file: Tables/CHAIR.tex
\begin{table*}[t]
\begin{center}
\begin{small}
\setlength{\tabcolsep}{8pt}
\caption{\textbf{CHAIR benchmark.}
Lower scores ($\downarrow$) on CHAIR$_S$, CHAIR$_I$ and higher ($\uparrow$) F1 indicate better performance. The best results in each metric are \textbf{bolded}, and the second-best are \underline{underlined}.}
\vspace{-0.1in}
\label{tab:CHAIR}
\resizebox{0.98\textwidth}{!}{
\begin{tabular}{lccccccccc}
    \toprule
      \multirow{2}{*}[-0.5ex]{\textbf{Method}}
        &  \multicolumn{3}{c}{\textbf{LLaVA-1.5-7B}}
        &  \multicolumn{3}{c}{\textbf{Qwen2.5-VL-7B}}
        &  \multicolumn{3}{c}{\textbf{InternVL3-8B}} \\
     \cmidrule(lr){2-4}\cmidrule(lr){5-7}\cmidrule(lr){8-10}
         & CHAIR$_S$ $\downarrow$ & CHAIR$_I$ $\downarrow$ & F1 $\uparrow$
         & CHAIR$_S$ $\downarrow$ & CHAIR$_I$ $\downarrow$ & F1 $\uparrow$
         & CHAIR$_S$ $\downarrow$ & CHAIR$_I$ $\downarrow$ & F1 $\uparrow$ \\
     \midrule
    Greedy & \tc{53.8} & \tc{14.66} & \tc{82.33} & \tc{35.2} & \tc{8.43} & \tc{79.85}
           & \tc{29.4} & \tb{7.91} & \tc{81.11} \\
    DoLa   & \tc{53.6} & \tc{14.45} & \tc{82.27} & \tc{31.0} & \tb{7.63} & \tc{79.33}
           & \tb{28.8} & \tc{8.49} & \tc{81.37} \\
    VCD    & \tc{52.8} & \tc{15.61} & \tc{81.84} & \tc{35.0} & \tc{9.07} & \tc{80.10}
           & \tc{30.8} & \tc{8.43} & \tc{81.56} \\
    M3ID   & \tc{57.0} & \tc{16.57} & \tc{82.06} & \tc{33.8} & \tc{9.80} & \tc{78.80}
           & \tc{29.2} & \tc{8.02} & \tc{81.81} \\
    CODE   & \ta{45.2} & \ta{13.13} & \tc{81.48} & \ta{17.0} & \tc{9.35} & \tc{48.50}
           & \tc{29.0} & \tc{8.25} & \tc{81.32} \\
    SID    & \tb{50.6} & \tc{13.58} & \ta{83.34} & \tc{37.6} & \tc{8.84} & \tb{80.40}
           & \tc{33.0} & \tc{9.58} & \tb{81.82} \\
    \ca{Ours} & \cb{50.6} & \cb{13.43} & \cb{83.17} & \cb{29.8} & \ca{7.04} & \ca{81.16}
              & \ca{27.6} & \ca{7.27} & \ca{82.29} \\
    \bottomrule
\end{tabular}
}
\end{small}
\end{center}
\vspace{-0.1in}
\end{table*}

%% file: Tables/POPE_summary.tex
\begin{table*}[t]
\begin{center}
\begin{small}
\setlength{\tabcolsep}{8pt}
\caption{
    \textbf{POPE benchmark}.
    Average for 3 datasets and 3 types. 
    Higher scores ($\uparrow$) on Accuracy, Precision, and F1 indicate better performance. 
    The best results in each metric are \textbf{bolded}, and the second-best are \underline{underlined}.
}
\vspace{-0.1in}
\label{tab:POPE_summary}
\resizebox{0.98\textwidth}{!}{
\begin{tabular}{lccccccccc}
    \toprule
      \multirow{2}{*}[-0.5ex]{\textbf{Method}}
        & \multicolumn{3}{c}{\textbf{LLaVA-1.5-7B}}
        & \multicolumn{3}{c}{\textbf{Qwen2.5-VL-7B}}
        & \multicolumn{3}{c}{\textbf{InternVL3-8B}} \\
     \cmidrule(lr){2-4}\cmidrule(lr){5-7}\cmidrule(lr){8-10}
         & Accuracy $\uparrow$ & Precision $\uparrow$ & F1 $\uparrow$
         & Accuracy $\uparrow$ & Precision $\uparrow$ & F1 $\uparrow$
         & Accuracy $\uparrow$ & Precision $\uparrow$ & F1 $\uparrow$ \\
     \midrule
    Greedy & \tc{79.47} & \tc{74.48} & \tc{82.36} & \tc{84.23} & \ta{93.01} & \tc{82.53}
           & \tb{89.17} & \tb{88.87} & \tb{89.36} \\
    DoLa   & \tc{79.58} & \tc{74.67} & \tc{82.41} & \tc{81.36} & \tc{92.36} & \tc{78.71}
           & \tc{86.63} & \ta{92.08} & \tc{85.64} \\
    VCD    & \tc{77.74} & \tc{72.51} & \tc{80.97} & \tc{84.43} & \tc{92.48} & \tc{82.91}
           & \tc{85.39} & \tc{86.33} & \tc{85.41} \\
    M3ID   & \tc{79.48} & \tc{74.48} & \tc{82.37} & \tc{83.56} & \tc{91.52} & \tc{82.00}
           & \tc{89.05} & \tb{88.87} & \tc{89.22} \\
    CODE   & \tc{78.58} & \tc{73.98} & \tc{81.35} & \tc{83.55} & \tc{90.20} & \tc{82.26}
           & \tc{86.70} & \tc{85.40} & \tc{87.22} \\
    SID    & \tb{80.40} & \tb{75.94} & \tb{82.91} & \ta{85.78} & \tc{92.27} & \ta{84.72}
           & \tc{85.38} & \tc{86.46} & \tc{85.44} \\
    \ca{Ours} & \ca{81.80} & \ca{78.03} & \ca{83.45} & \cb{84.53} & \cb{93.00} & \cb{82.93}
              & \ca{89.21} & \cc{88.63} & \ca{89.44} \\
    \bottomrule
\end{tabular}
}
\end{small}
\end{center}
\vspace{-10pt}
\end{table*}

%% file: Tables/VQA.tex
\begin{table*}[t]
\begin{center}
\begin{small}
\setlength{\tabcolsep}{8pt}
\caption{
    \textbf{VQAv2 and MMMU}. 
    We report \textit{Accuracy} for both benchmarks. 
    Higher scores ($\uparrow$) indicate better performance. 
    The best results in each setting are \textbf{bolded}, and the second-best are \underline{underlined}.
}
\vspace{-0.1in}
\label{tab:mmmu_vqav2}
\resizebox{0.95\textwidth}{!}{
\begin{tabular}{lccc|ccc}
    \toprule
      \multirow{2}{*}[-0.5ex]{\textbf{Method}} 
        & \multicolumn{3}{c}{\textbf{VQAv2} (Accuracy $\uparrow$)} 
        & \multicolumn{3}{c}{\textbf{MMMU} (Accuracy $\uparrow$)} \\
      \cmidrule(lr){2-4}\cmidrule(lr){5-7}
        & \textbf{LLaVA-1.5-7B} & \textbf{Qwen2.5-VL-7B} & \textbf{InternVL3-8B}
        & \textbf{LLaVA-1.5-7B} & \textbf{Qwen2.5-VL-7B} & \textbf{InternVL3-8B} \\
     \midrule
    Greedy 
      & \tc{69.47} & \tc{65.80} & \tb{68.73}
      & \tc{34.39} & \tc{51.11} & \tb{53.54} \\
    DoLa   
      & \tc{69.87} & \tc{60.60} & \tc{68.13}
      & \tc{34.29} & \tc{50.15} & \tc{53.08} \\
    VCD    
      & \tc{65.93} & \tc{66.07} & \tc{53.60}
      & \tc{33.51} & \tb{51.13} & \tc{52.60} \\
    M3ID   
      & \tb{70.33} & \tc{59.87} & \ta{69.53}
      & \tb{34.51} & \tc{46.63} & \tc{53.43} \\
    CODE   
      & \tc{66.00} & \tc{46.60} & \tc{56.33}
      & \tc{34.48} & \tc{42.79} & \tc{52.41} \\
    SID    
      & \tb{70.33} & \tb{67.13} & \tc{55.27}
      & \tc{33.10} & \tc{49.75} & \tc{51.85} \\
    \ca{Ours} 
      & \ca{70.40} & \ca{67.60} & \cc{68.40}
      & \ca{35.13} & \ca{51.18} & \ca{54.14} \\
    \bottomrule
\end{tabular}
}
\end{small}
\end{center}
\vspace{-0.1in}
\end{table*}

%% file: Tex/Conclusion.tex
\section{Conclusion}
\label{sec:conclusion}

In this paper, we revisited the role of vision tokens in LVLM decoding and provided quantitative evidence that they encode \emph{interpretable} semantics that become explicit under semantically coherent vocabulary constraints. Building on these findings, we introduced \name{}, a training-free, model-agnostic decoding strategy that references the most relevant vision token via a \emph{context-aware constrained divergence} and refines token logits accordingly. 
As demonstrated by extensive experiments, \name{} consistently improves performance across various state-of-the-art LVLMs, demonstrating scalability and broad applicability without architectural modification.

%% file: Tex/Limitation_Ethical.tex
\section*{Limitations}
A potential limitation of our method is its strong reliance on the semantics encoded in vision tokens, which may lead to overfitting to visually salient cues.
While this behavior improves visual grounding in most cases, it can result in an overemphasis on perceptually grounded elements at the expense of implicit, commonsense, or non-visual contextual information. 
Although we compare fixed interpolation schemes in Appendix~E.5, more adaptive interpolation or confidence-calibration strategies between the reference and vanilla logits remain worth exploring, as a context-dependent fusion rule may better balance visual grounding against non-visual reasoning demands.

Another limitation is its exclusive reliance on the model’s internal representations without access to external knowledge sources. 
As a result, it cannot correct factual errors or hallucinations originating from pretraining data, particularly in cases where visual evidence alone is insufficient for disambiguation or grounding. 
While our method improves the alignment between visual inputs and generated text, it does not incorporate retrieval mechanisms that could supplement the model with up-to-date or context-specific information. 
Nevertheless, due to its modular and decoding-time nature, our method is compatible with retrieval-augmented generation (RAG) pipelines and could be integrated with external knowledge modules in future work.

Finally, our empirical analysis and experiments are conducted on a limited set of open-source LVLMs in the 7--32B parameter range.
Due to computational resource constraints, we did not evaluate our method on 70B-scale or larger models, and we therefore do not claim that the observed gains directly transfer to substantially larger or proprietary LVLMs.

\section*{Ethical Considerations}
This work relies on publicly available multimodal benchmarks such as HallusionBench, CHAIR, POPE, VQAv2, and MMMU, and on released LVLM checkpoints including LLaVA-1.5-7B, Qwen2.5-VL-7B, and InternVL3-8B.
We do not collect new data or interact with human subjects, and we follow the original licensing and usage conditions of all datasets and models.

As with any approach that operates on LVLMs, our method inherits any biases, uneven coverage, or other failure modes present in the underlying models and training corpora.
Since our contribution targets the decoding procedure by improving visual grounding and reducing object hallucinations, it does not introduce new training data or objectives explicitly aimed at mitigating social biases.
Potential deployments of LVLMs enhanced with our method, especially in high-stakes or user-facing applications, should therefore include appropriate human oversight and domain-specific safeguards.

\section*{Acknowledgments}

All authors are affiliated with the Department of Artificial Intelligence at Yonsei University.
This research was supported in part by Institute for Information \& communications Technology Planning \& Evaluation (IITP) grant funded by the Korea government (MSIT) (No. RS-2020-II201361, Artificial Intelligence Graduate School Program (Yonsei University); No. RS-2025-02215344, Development of AI Technology with Robust and Flexible Resilience Against Risk Factors; No. RS-2025-25442405, Development of a Self-Learning World Model-Based
AGI System for Hyperspectral Imaging).

%% file: Tex/Appendix.tex
\clearpage
\appendix

\section{Additional Related Works}
\label{appendix:extended-related-work}

In this section, we provide an extended discussion of the literature on aligning visual inputs with language generation in LVLMs, expanding upon the taxonomy introduced in Sec.~\ref{sec:related}.
While our primary classification follows the temporal stage of intervention---\textit{pre-alignment}, \textit{intra-alignment}, and \textit{post-alignment}---we also highlight how interpretability-driven projection methods have recently been integrated into these frameworks.

\paragraph{Pre-alignment.}
This line of research focuses on establishing foundational cross-modal mapping during the model's development phase. 
Recent state-of-the-art LVLMs refine these mappings by coupling decoder-only backbones with high-resolution vision encoders~\citep{Qwen-VL,chen2024internvl} or introducing dynamic resolution processing to accommodate diverse aspect ratios~\citep{bai2025qwen2}. 
Other efforts improve fine-grained spatial grounding via hybrid region representations~\citep{you2023ferret} or probabilistic vision tokenization schemes designed to better match language embeddings~\citep{lu2024ovis}. 
While these methods enhance the model's inherent capacity, they are static and cannot dynamically correct errors that emerge during the inference process.

\paragraph{Intra-alignment.}
Methods in this category intervene during the decoding process itself to guide the model toward visually grounded outputs. 
Perturbation-based approaches contrastively compare logits under original and perturbed inputs (e.g., noisy images or instruction disturbances) to suppress hallucinations caused by language priors~\citep{leng2024vcd,wang2024icd,favero2024m3id,kim2024code}.
Calibration-based methods instead focus on the internal attention mechanism; they recalibrate how the decoder attends to or weights visual tokens by deriving vision-aware adjustments from learned attention patterns~\citep{gong2024damro,liu2024pai,woo2024avisc,huo2024sid,kang2025see}.
Notably, \citet{jiang2025devils} further explore this by diagnosing and editing attention patterns in middle layers to mitigate unwanted artifacts during inference.
A third family introduces auxiliary visual-sensitive distributions or penalty terms to directly modify decoding rules so that tokens are more strongly grounded in the visual input~\citep{chen2024halc,huang2024opera,wang2024deco}.
Across these lines of work, visual information is typically incorporated implicitly through hidden states, logits, or attention scores, leaving the semantic role of individual vision tokens latent.
In contrast, \name{} makes the textual semantics of individual vision tokens explicit in the language space and leverages their induced distributions as a context-aware constraint, providing an interpretable and effective handle for steering LVLM decoding.

\paragraph{Post-alignment.}
This stage operates on the generated sequence to identify and correct hallucinations after the decoding is complete.
Traditional methods use auxiliary revisors or multi-stage verification pipelines conditioned on external VQA models~\citep{zhou2023lure,yin2024woodpecker}. 
Recently, interpretability-based techniques have been adopted for diagnostic purpose; for instance, \citet{jiang2025interpreting} project vision tokens into the language space to diagnose discrepancies between visual perception and generated text as a post-hoc verification signal.
While such methods provide a diagnostic signal for post-hoc verification, \name{} utilizes similar projection techniques as a \textit{predictive} signal during decoding to prevent hallucinations before they are fully generated.

\clearpage
\section{Additional Motivation and Analyses}
\label{appendix:motivation-details}

\subsection{Metric formulation: \emph{at-least-one recall@}k}
\label{appendix:metric-def}
\paragraph{Definition.}

Given a hidden state $h$ and the corresponding text token distribution 
$p_{\theta}(h,\mathcal{V})$ over the vocabulary $\mathcal{V}$ 
as defined in Eq.~\eqref{eq:softmax}, 
we evaluate whether at least one ground-truth (GT) object synonym is ranked 
within the top-$k$ candidates.

Let $\mathcal{U}$ denote a collection of evaluation units 
(e.g., hallucinated decoding steps or vision tokens, depending on the analysis).
For each $u\in\mathcal{U}$, let $h^{(u)}$ be the associated hidden state, 
and let $G_u\subseteq\mathcal{V}$ be the set of GT object synonyms relevant to $u$.

We define the top-$k$ set
\[
\mathrm{TopK}_k\!\bigl(p_{\theta}(h^{(u)},\mathcal{V})\bigr)
\subseteq\mathcal{V}
\]
as the set of $k$ tokens with the highest probabilities under 
$p_{\theta}(h^{(u)},\mathcal{V})$.

Then, at-least-one recall@$k$ over $\mathcal{U}$ with evaluation vocabulary
$\mathcal{V}$ is defined as

\begin{equation}
    \label{eq:aleastone_mean}
    \resizebox{\columnwidth}{!}{$
    \begin{aligned}
        &\text{at-least-one recall@}k(\mathcal{U};\mathcal{V}) \\
        &=\;
        \frac{1}{|\mathcal{U}|}
        \sum_{u\in\mathcal{U}}
        \mathbf{1}\!\left\{
            \mathrm{TopK}_k\!\bigl(p_{\theta}(h^{(u)},\mathcal{V})\bigr)
            \;\cap\;G_u\;\neq\;\varnothing
        \right\}.
    \end{aligned}
    $}
\end{equation}

\subsection{Step-level hallucination analysis for Figure~\ref{fig:motivation_exp1}}
\label{appendix:exp1}
\paragraph{Setup formulation.}
CHAIR~\citep{rohrbach2018chair} evaluates captions against MS-COCO object annotations
under an object-existence criterion.
On the 500 captions generated in default greedy setup, CHAIR identifies 275 hallucinated words 
in total.
We then move to step-level analysis, where each hallucinated object mention
defines a decoding step (i.e., an evaluation unit).

Among these 275 hallucinated steps, we retain only those in which 
(i) the hallucinated word and 
(ii) at least one ground-truth (GT) object synonym for the same image 
are both tokenized as a single token by the tokenizer.
This restriction allows each retained object label to be aligned directly with a specific token index, yielding $n=190$ hallucinated decoding steps for step-level analysis.

In the notation of Appendix~\ref{appendix:metric-def}, we instantiate
$\mathcal{U}$ as this set of hallucinated decoding steps and $\mathcal{V}$ as
the full vocabulary, and compute
$\text{at-least-one recall@}k(\mathcal{U};\mathcal{V})$ with $|\mathcal{U}| = 190$.

\paragraph{Reporting.}
Table~\ref{tab:ci} reports the resulting recall values with 95\% Wilson score
intervals.

\begin{table}[h]
    \small
    \caption{\textbf{Step-level GT recall at hallucinated positions.} At-least-one GT recall at hallucinated steps ($n=190$) with 95\% Wilson intervals.}
    \label{tab:ci}
    \vspace{0.2cm}
    \begin{tabular}{lrc}
        \hline
        Top-$k$ & \% of vocab & \emph{at-least-one recall@}k [95\% CI] \\
        \hline
        Top-50  & 0.033\% & 63.2\% [56.0, 70.0] \\
        Top-100 & 0.066\% & 73.7\% [66.9, 79.6] \\
        Top-500 & 0.33\%  & 95.8\% [91.6, 98.1] \\
        \hline
    \end{tabular}
\end{table}

\subsection{Paired GT vs. hallucinated probabilities for Table~\ref{tab:motivation_exp2}}
\label{appendix:exp2}

\paragraph{Setup formulation.}
Given an output hidden state $h_{T+t-1}^L$ from the last decoder layer $L$ at
decoding step $t$ and the corresponding text token distribution 
$p_{\theta}(h_{T+t-1}^L,\mathcal{V})$ over the vocabulary $\mathcal{V}$ 
as defined in Eq.~\eqref{eq:softmax}, we compare how much probability mass is
assigned to ground-truth versus hallucinated objects.

For CHAIR, let $\mathcal{V}_{\texttt{coco}}$ denote the MS-COCO synonym set ($|\mathcal{V}_{\texttt{coco}}|=403$).
To obtain unambiguous token–object correspondence and avoid multi-token
variants, we construct the single-token subset
\begin{equation}
    \resizebox{\columnwidth}{!}{$
        \mathcal{V}_{\texttt{obj}}
        \;=\;
        \bigl\{\, w\in\mathcal{V}_{\texttt{coco}}
        \;:\;
        \texttt{len}(\texttt{tokenize}(\texttt{"\textvisiblespace{}}w\texttt{"})) = 1
        \,\bigr\},
    $}
    \label{eq:176subset}
\end{equation}
which contains 176 word types covering 62 object categories.

At decoding position $t$, we define the CHAIR-based ground-truth and
hallucinated synonym subsets over $\mathcal{V}_{\texttt{obj}}$ as
$\mathcal{G}_t\subseteq\mathcal{V}_{\texttt{obj}}$ and
$\mathcal{H}_t=\mathcal{V}_{\texttt{obj}}\setminus\mathcal{G}_t$, respectively.
For step-level analyses, we further require that
(i) the hallucinated word at $t$ is single-tokenizable and
(ii) $\mathcal{G}_t\neq\varnothing$, yielding $n=190$ hallucinated decoding
steps (selection details match Appendix~\ref{appendix:exp1}).
We evaluate two conditions:
\emph{w/ image} (vision tokens present) and
\emph{w/o image} (all vision tokens removed); in the latter, decoding positions
are aligned by teacher forcing with the previously generated caption while
keeping all other decoding details identical.

\begin{table*}[t]
    \centering
    \caption{\textbf{Wilcoxon signed-rank comparison of GT vs.\ hallucinated scores.}
    $\Delta_t=(\mathrm{GT}-\mathrm{Hal})_{\mathrm{with}}
              -(\mathrm{GT}-\mathrm{Hal})_{\mathrm{w/o}}$.
    $N$ is the number of paired decoding positions, and
    Holm-adjusted $p$-values are reported for $\Delta_t$.}
    \label{tab:wilcoxon}
    \vspace{0.15cm}
    \resizebox{1.0\textwidth}{!}{
    \begin{tabular}{lrrrrr}
        \toprule
        Setting & $N$ & median$(d^{\mathrm{with}})$ & median$(d^{\mathrm{w/o}})$ & median$(\Delta)$ & $p_{\text{Holm}}$ on $\Delta$ \\
        \midrule
        All tokens & 88{,}706 & $-3.289$ & $-4.826$ & $\;\;0.715$ & $<10^{-300}$ \\
        All tokens (in hallucinated set) & 35{,}116 & $-3.720$ & $-5.273$ & $\;\;0.774$ & $1.09{\times}10^{-116}$ \\
        Hallucinated steps & 190 & $-8.281$ & $-9.435$ & $-0.180$ & $0.95$ \\
        \bottomrule
    \end{tabular}}
\end{table*}

For each decoding position $t$, we operate directly on the object-only
distribution $p_{\theta}(h_{T+t-1}^L,\mathcal{V}_{\texttt{obj}})$.
Let
\[
\begin{aligned}
    m_t^{\mathrm{GT}}
    &= \max_{w\in\mathcal{G}_t} p_{\theta}(w\mid h_{T+t-1}^L,\mathcal{V}_\texttt{obj}), \\
    m_t^{\mathrm{Hal}}
    &= \max_{w\in\mathcal{H}_t} p_{\theta}(w\mid h_{T+t-1}^L,\mathcal{V}_\texttt{obj}),
\end{aligned}
\]
and define the corresponding quantities $m_t^{\mathrm{GT,w/o}}$ and
$m_t^{\mathrm{Hal,w/o}}$ under the w/o-image condition.
Table~\ref{tab:motivation_exp2} reports the averages of these per-step maxima of object probabilities.

\paragraph{Statistical test and reporting.}
For paired comparisons, we form the step-level differences
\[
\begin{aligned}
    d_t^{\mathrm{with}} &= m_t^{\mathrm{GT}} - m_t^{\mathrm{Hal}},\\
    d_t^{\mathrm{w/o}}  &= m_t^{\mathrm{GT,w/o}} - m_t^{\mathrm{Hal,w/o}},\\
    \Delta_t &= d_t^{\mathrm{with}} - d_t^{\mathrm{w/o}}.
\end{aligned}
\]
We apply the Wilcoxon signed-rank test (two-sided; Pratt) to
$d_t^{\mathrm{with}}$, $d_t^{\mathrm{w/o}}$, and $\Delta_t$ on the
intersection of decoding positions available in both conditions.
We report the paired sample size $N$, the median of each quantity, and
Holm-adjusted $p$-values for $\Delta_t$.

In the full-token analyses, the median $\Delta$ is positive and highly
significant, indicating a systematic shift in the step-level difference
$m_t^{\mathrm{GT}} - m_t^{\mathrm{Hal}}$ toward GT when vision tokens are
present.
For the hallucinated-step subset ($n{=}190$), the median $\Delta$ is small
and not significant, i.e., the reversed ordering (Hal $>$ GT) persists at
those positions despite the image.

\subsection{Vision-token projection analysis for Figure~\ref{fig:motivation_exp3}}
\label{appendix:exp3}

\paragraph{Setup formulation.}
We focus on the final decoder layer ($L$) and treat each vision token as an
evaluation unit.
Let the 500 MS-COCO validation images be indexed by
$n \in \{1,\dots,500\}$.
For image $n$, let
$\mathbf{v}^{(n)} = (v^{(n)}_1,\dots,v^{(n)}_{|\mathbf{v}^{(n)}|})$
denote the sequence of vision tokens, and let
$h^{L,(n)}_i$ be the corresponding final-layer hidden state of
$v^{(n)}_i$.

We define the evaluation set $\mathcal{U}$ as the collection of all vision
tokens across the 500 images,
\[
\resizebox{1.0\columnwidth}{!}{$
    \mathcal{U}
    \;=\;
    \left\{ (n,i) \;:\; n \in \{1,\dots,500\},\;
                       i \in \{1,\dots,|\mathbf{v}^{(n)}|\} \right\},
$}
\]
and denote its size by
\[
    |\mathcal{U}| \;=\; N \;=\; \sum_{n=1}^{500} |\mathbf{v}^{(n)}|
    \;=\; 173{,}173.
\]
For each unit $u = (n,i) \in \mathcal{U}$, we write the associated hidden
state as $h^{(u)} = h^{L,(n)}_i$.

Passing $h^{(u)}$ through the language-modeling head yields a text token
distribution over the full vocabulary $\mathcal{V}$,
$p_{\theta}(h^{(u)},\mathcal{V})$.
Additionally, we reuse $\mathcal{V}_{\texttt{obj}}$, the single-token MS-COCO
synonym subset defined in Eq.~\eqref{eq:176subset} (176 word types covering
62 object categories), and obtain $p_{\theta}(h^{(u)},\mathcal{V}_{\texttt{obj}})$
by applying softmax over $\mathcal{V}_{\texttt{obj}}$.

For each vision-token unit $u \in \mathcal{U}$, we construct a GT synonym set
$G_u$ from the MS-COCO categories annotated in the same image as $h^{(u)}$,
restricted to the 62 categories covered by $\mathcal{V}_{\texttt{obj}}$.
In the notation of Appendix~\ref{appendix:metric-def}, we thus instantiate
$\mathcal{U}$ as the set of all $N$ vision tokens and consider two evaluation
vocabularies, $\mathcal{V}$ and $\mathcal{V}_{\texttt{obj}}$.
We then compute
$
\text{at-least-one recall@}k(\mathcal{U};\mathcal{V})
$
and
$
\text{at-least-one recall@}k(\mathcal{U};\mathcal{V}_{\texttt{obj}})
$
according to Eq.~\eqref{eq:aleastone_mean}.

\paragraph{Reporting.}
Table~\ref{tab:exp3} summarizes token-level
at-least-one recall@$k(\mathcal{U};\mathcal{V})$ and
at-least-one recall@$k(\mathcal{U};\mathcal{V}_{\texttt{obj}})$
over $|\mathcal{U}| = N = 173{,}173$ vision-token projections, together with
95\% Wilson score intervals.

\begin{table}[h]
    \caption{\textbf{Vision-token at-least-one GT recall.}
    At-least-one GT recall over $N{=}173{,}173$ vision-token projections
    with 95\% Wilson score intervals. GT targets are restricted to the same 62
    categories in both settings.}
    \label{tab:exp3}
    \vspace{3pt}
    \setlength{\tabcolsep}{8pt}
    \resizebox{1.0\columnwidth}{!}{
    \begin{tabular}{lcc}
        \toprule
        Top-$k$ & $\text{at-least-one recall@}k(\mathcal{V})$ & $\text{at-least-one recall@}k(\mathcal{V}_\texttt{obj})$ \\
        \midrule
        Top-1  & 2.03\% [1.96, 2.10] & 40.44\% [40.21, 40.67] \\
        Top-5  & 7.87\% [7.74, 8.00] & 69.56\% [69.34, 69.78] \\
        Top-10 & 12.38\% [12.23, 12.54] & 79.78\% [79.59, 79.97] \\
        Top-30 & 21.72\% [21.53, 21.91] & 89.24\% [89.09, 89.39] \\
        \bottomrule
    \end{tabular}
    }
\end{table}

\newpage
\section{Additional Method Details: \name{} Decoding}
\label{appendix:revisit}

\subsection{Step-by-step decoding overview with Figure~\ref{fig:overview}}
\label{appendix:revisit_overview}

To complement the formal description in Sec.~\ref{sec:method}, 
we present a step-by-step overview of \name{} and its pseudo-code implementation.
Fig.~\ref{fig:overview} illustrates a single decoding step on a representative example, 
while Alg.~\ref{alg:main} summarizes the full procedure over the entire decoding trajectory.
\name{} operates purely at decoding time: the LVLM architecture and parameters are kept fixed, and only the decoding function is modified.
By caching vision-token logits once and manipulating them through slicing and masking, 
our implementation is memory-efficient, model-agnostic, and requires no additional forward passes or training.

At a high level, \name{} modifies the vanilla next-token distribution only when vision tokens provide a more semantically consistent signal.
Given the fruit image and the textual query ``\texttt{How many uncut fruits are in the image?}'' from the LLaVA-Bench-In-the-Wild dataset~\citep{liu2023visual}, 
the LVLM produces at each decoding step $t$ a vanilla distribution 
$p_{\theta}(h_{T+t-1}^L, \mathcal{V})$ over the full vocabulary (Eq.~\eqref{eq:softmax}).
At $t=0$ and $t=1$, the model generates ``\texttt{There}'' and ``\texttt{are}'', and \name{} leaves these predictions unchanged because the vanilla distribution is already highly confident.
At $t=2$, however, the vanilla distribution erroneously favors ``\texttt{four}'' over the correct answer ``\texttt{three}''.

Concretely, in step (1), the vanilla distribution is restricted to a context-aware constrained subset 
$\mathcal{V}_{\texttt{cons}}^t$ via the adaptive plausibility constraint in Eq.~\eqref{eq:our_apc}.
All subsequent operations are performed on this subset, and tokens outside $\mathcal{V}_{\texttt{cons}}^t$ are effectively removed from consideration.

In step (2), for each candidate vision token hidden state $h_i^j$ from layer $j \in \mathcal{J}$,
we obtain its projected distribution over $\mathcal{V}_{\texttt{cons}}^t$,
$p_{\theta}(h_i^j,\mathcal{V}_{\texttt{cons}}^t)$, as defined in Eq.~\eqref{eq:vision_project}.
We then measure the Jensen--Shannon divergence between
$p_{\theta}(h_{T+t-1}^L,\mathcal{V}_{\texttt{cons}}^t)$ and each
$p_{\theta}(h_i^j,\mathcal{V}_{\texttt{cons}}^t)$,
and select the pair $(i^*, j^*)$ with the smallest divergence as the most relevant vision token for the current context.

In step (3), the vanilla distribution and the selected vision-token distribution are fused.
For tokens in $\mathcal{V}_{\texttt{cons}}^t$, the final distribution $p_{\tt fin}$ is obtained via log-space addition (Eq.~\eqref{eq:log-space-addition}) followed by normalization over the constrained set (Eq.~\eqref{eq:final_fusion});
for tokens outside $\mathcal{V}_{\texttt{cons}}^t$, the probability mass is set to zero.
The next token $y_t$ is then sampled from $p_{\tt fin}$.
In the example of Fig.~\ref{fig:overview}, this refinement step shifts the decision from ``\texttt{four}'' to the visually grounded answer ``\texttt{three}''.
Appendix~\ref{appendix:motivation-fig} provides the exact token-level probabilities underlying this example, together with additional case studies.

\subsection{Theoretical view of logit refinement}
\label{appendix:revisit_theory}
The refinement rule in Sec.~\ref{sec:method_logit_refinement} (Eq.~\eqref{eq:log-space-addition}) admits a simple KL-regularized interpretation.
Let \(p_{\text{output}}(y)\) denote the vanilla next-token distribution on the constrained candidate set \(\mathcal{V}^t_{\texttt{cons}}\), and let \(p_{\text{vision}}(y)\) denote the selected vision-token distribution on the same set.
We consider the objective
\[
q^* = \arg\max_{q \in \Delta} \; \mathbb{E}_{y\sim q}[\log p_{\text{vision}}(y)] - \gamma\,\mathrm{KL}(q \,\|\, p_{\text{output}}),
\]
which encourages the refined distribution to assign higher mass to tokens preferred by the vision-induced reference while remaining close to the model's original distribution.
This objective yields the closed-form solution
\[
q^*(y) \propto p_{\text{output}}(y)\, p_{\text{vision}}(y)^{1/\gamma}.
\]
Hence, the proposed refinement is equivalent to a Product-of-Experts (PoE)-style combination in probability space, or additive fusion in log space.
The default rule in the main text corresponds to the equal-strength case (i.e., $\gamma = 1$).

\onecolumn
\input{Tex/Algorithm}

\clearpage
\twocolumn
\section{Additional Experimental Details}
\subsection{Dataset and evaluation setup}
\label{appendix:dataset-details}

\paragraph{HallusionBench evaluation.}
HallusionBench~\citep{guan2024hallusionbench} is a diagnostic benchmark for image-context reasoning that disentangles \emph{Language Hallucination} (over-reliance on parametric priors) and \emph{Visual Illusion} (misinterpretation of visual inputs).
All questions are binary (Yes/No) and organized into \emph{control pairs} to test consistency across visual contexts.
We focus on the \textit{Visual Dependent (VD)} split, which comprises 591 Yes/No questions that \emph{require} the image to answer; each figure consists of an original image and one or more human-edited variants (e.g., flipping, order reversing, masking, optical-character editing, object editing, color editing) to stress visual robustness.
We use the official data and evaluation scripts.\footnote{\url{https://github.com/tianyi-lab/HallusionBench}}

\textit{Protocol.}
Following the official setup, we adopt the \emph{text-only GPT-4-assisted} judging scheme.
Let figures be indexed by $i \in \mathbb{I}$.
For each figure $i$, let $\{I_{i,j}\}_{j \in \mathcal{J}_i}$ denote its image variants with $j=0$ the original and $j \ge 1$ edited versions, and let $\{q_{i,k}\}_{k \in \mathcal{K}_i}$ be the associated Yes/No questions (shared across variants).
Given a model’s raw response $\mathcal{M}(I_{i,j}, q_{i,k})$, the judge (GPT-4, temperature 0) receives a textual prompt comprising \{$q_{i,k}$ (question), $y_{i,j,k}$ (reference answer), $\mathcal{M}(I_{i,j}, q_{i,k})$ (model response)\} and returns a label in \{\texttt{Correct}, \texttt{Incorrect}, \texttt{Uncertain}\}.
We binarize correctness as
\[
\resizebox{1.0\columnwidth}{!}{$
    \begin{aligned}
        &b_{\mathcal{M}}(I_{i,j}, q_{i,k}) \\
        &=\;
        \begin{cases}
            1, & \text{if the judge returns \texttt{Correct}},\\
            0, & \text{if the judge returns \texttt{Incorrect} or \texttt{Uncertain}}.
        \end{cases}
    \end{aligned}
$}
\]

\textit{Metrics.}
Let $\mathcal{T}=\{(i,j,k): i \in \mathbb{I},\, j \in \mathcal{J}_i,\, k \in \mathcal{K}_i\}$ be all VD image–question triples,
and let $\mathbb{Q}=\{(i,k): i \in \mathbb{I},\, k \in \mathcal{K}_i\}$ be the set of control \emph{question pairs}.
We report the standard HallusionBench accuracies:
\[
\resizebox{1.0\columnwidth}{!}{$
    \begin{aligned}
        \text{aAcc} \;&=\; \frac{1}{|\mathcal{T}|} \sum_{(i,j,k)\in\mathcal{T}} b_{\mathcal{M}}(I_{i,j}, q_{i,k}), \\
        \text{qAcc} \;&=\; \frac{1}{|\mathbb{Q}|} \sum_{(i,k)\in\mathbb{Q}} \mathbf{1}\!\Big\{\,\bigwedge_{j\in\mathcal{J}_i} b_{\mathcal{M}}(I_{i,j}, q_{i,k}) = 1 \,\Big\}, \\
        \text{fAcc} \;&=\;
        \frac{1}{\sum_{i \in \mathbb{I}} |\mathcal{J}_i|}
        \sum_{i \in \mathbb{I}} \sum_{j \in \mathcal{J}_i}
        \mathbf{1}\!\Big\{\,\bigwedge_{k \in \mathcal{K}_i} b_{\mathcal{M}}(I_{i,j}, q_{i,k}) = 1 \,\Big\}.
    \end{aligned}
$}
\]

Here, \textbf{aAcc} is \emph{accuracy per question} (overall), \textbf{qAcc} is \emph{accuracy per question pair} (the same question must be correct across the figure’s image variants), and \textbf{fAcc} is \emph{accuracy per image} (all questions assigned to a given image—original or edited—must be correct).
Following the leaderboard convention, we additionally report \textbf{Easy aAcc} (restricted to original images, $j{=}0$) and \textbf{Hard aAcc} (restricted to edited images, $j{\ge}1$).

\paragraph{CHAIR evaluation.}
CHAIR~\citep{rohrbach2018chair} is a generative benchmark designed to measure object hallucinations in image captioning. 
A hallucinated object refers to any entity mentioned in the generated caption that is not present in the corresponding image. 
This metric is widely adopted to evaluate the visual grounding capability of vision-language models.

\textit{Protocol.}
We follow the standard CHAIR evaluation procedure based on the MS-COCO annotations and the official implementation.\footnote{\url{https://github.com/LisaAnne/Hallucination}} 
Each image is associated with a set of ground-truth objects defined by the 80 MS-COCO detection categories. 
During evaluation, noun phrases extracted from generated captions are matched against this predefined object list; mentions of out-of-vocabulary objects are ignored, while in-vocabulary objects absent from the ground-truth annotations are considered hallucinated.
We evaluate on 500 images randomly sampled from the COCO validation set with a fixed seed (seed$=42$) for reproducibility, prompting each image with ``\texttt{Please describe this image in detail.}''.

\textit{Metrics.}
We report the two standard CHAIR metrics and an additional \textbf{F1}:
\begin{equation*}
\resizebox{1.0\columnwidth}{!}{$
    \begin{aligned}
        \text{CHAIR}_{S} &= \frac{\left|\{\text{hallucinated captions}\}\right|}{\left|\{\text{total captions}\}\right|}, \\
        \text{CHAIR}_{I} &= \frac{\left|\{\text{hallucinated object mentions}\}\right|}{\left|\{\text{total object mentions}\}\right|}, \\
        \text{F1} &= \frac{\sum_{\text{images}} \text{F1}_{\text{sample}}}{\left|\{\text{total captions}\}\right|}. \\
        \text{where}\quad
        &\text{F1}_{\text{sample}} =
        \begin{cases}
            \displaystyle \frac{2\,\text{Prec}\cdot\text{Rec}}{\text{Prec}+\text{Rec}}, & \text{if } \text{Prec}+\text{Rec}>0,\\[6pt]
            0, & \text{otherwise,}
        \end{cases} \\[6pt]
        \text{with}\quad
        &\text{Prec} = \frac{\left|\{\,\text{predicted objects that are in GT}\,\}\right|}{\left|\{\text{predicted objects}\}\right|}\;\; \\
        &\qquad \text{(duplicates in predictions counted)}, \\
        &\;\text{Rec}  = \frac{\left|\{\,\text{GT objects mentioned at least once}\,\}\right|}{\left|\{\text{GT objects}\}\right|}\;\; \\
        &\qquad \text{(GT is a set; no duplicates).}
    \end{aligned}
$}
\end{equation*}

\paragraph{POPE evaluation.}
POPE~\citep{li2023pope} is a discriminative benchmark designed to evaluate the object-level visual grounding capabilities of LVLMs. 

\textit{Protocol.}
It formulates the task as a binary object presence question-answering problem, where the model is asked to determine whether a specific object is present in a given image.
The benchmark covers three datasets: MS-COCO~\citep{lin2014mscoco}, A-OKVQA~\citep{schwenk2022okvqa}, and GQA~\citep{hudson2019gqa}. 
For each dataset, POPE defines three types of query strategies—\textit{random}, \textit{popular}, and \textit{adversarial}—yielding a total of nine evaluation scenarios. 
Binary questions are constructed in the form ``\texttt{Is there a <object> in the image?}'', with the queried object selected according to the designated strategy. 
Random queries are uniformly sampled from the dataset's object vocabulary. 
Popular queries focus on frequently occurring objects, while adversarial queries target objects that are semantically plausible in the given context but are absent from the image, thus probing the model’s reliance on prior co-occurrence statistics rather than visual evidence.
Ground-truth labels indicating object presence are provided by the official POPE benchmark.\footnote{\url{https://github.com/AoiDragon/POPE}} 
For MS-COCO, labels are derived from object detection annotations, whereas for A-OKVQA and GQA, they are obtained via SEEM-based segmentation annotations. 
Each query type consists of 3,000 binary QA instances, evenly split between positive and negative samples, resulting in a total of 27,000 QA examples.

\textit{Metrics.}
Evaluation is conducted using standard classification metrics derived from the confusion matrix, including accuracy, precision, and F1 score. 
Accuracy measures the overall proportion of correct predictions, while precision captures the fraction of predicted positive instances that are indeed correct.
F1 score, defined as the harmonic mean of precision and recall, provides a balanced assessment of model performance on this binary classification task.

\paragraph{VQAv2 evaluation.}
VQAv2~\citep{goyal2017vqav2} is a large-scale open-ended visual question answering benchmark where each question is annotated with 10 human answers.

\textit{Protocol.}
We use the official evaluation protocol and code for computing agreement with the 10 human annotations.\footnote{\url{https://visualqa.org/evaluation.html}} 
Since our models are prompted in a free-form manner and may generate short phrases or sentences rather than a single word, 
we first tokenize each generated answer into a sequence of whitespace-separated tokens $(w_1,\dots,w_T)$.
For each token $w_t$, we query the official VQAv2 scoring function $\text{Acc}(w_t)$, and define the per-question score as the maximum token-level score $\max_t \text{Acc}(w_t)$.
This procedure allows us to fairly evaluate free-form answers while remaining compatible with the official scoring rule.

\textit{Metrics.}
Concretely, for any candidate answer string $\textit{ans}$, the VQAv2 accuracy is
\begin{equation*}
\resizebox{1.0\columnwidth}{!}{$
    \text{Acc}(\textit{ans}) \;=\; \min\!\left\{\frac{\#\text{ of humans that said }\textit{ans}}{3},\, 1\right\},
$}
\end{equation*}
where the numerator counts how many of the 10 annotators produced $\textit{ans}$.
Given a model prediction tokenized as $(w_1,\dots,w_T)$, the per-question accuracy used in our experiments is
\begin{equation*}
    \text{Acc}_{\text{question}} \;=\; \max_{t \in \{1,\dots,T\}} \text{Acc}(w_t),
\end{equation*}
and the dataset accuracy is the mean of $\text{Acc}_{\text{question}}$ over all questions.

\paragraph{MMMU evaluation.}
MMMU~\citep{vqa_yue2024mmmu} is a large-scale, multi-discipline VQA benchmark that evaluates college-level multimodal understanding and reasoning across six disciplines, 30 subjects, and heterogeneous image types (charts, diagrams, tables, maps, music sheets, chemical structures, etc.). We use the validation split for controlled comparisons.

\textit{Protocol.}
The validation split includes open-ended and multiple-choice (MC) questions. Since our LLaVA-1.5 configuration supports only single-image inputs, we exclude the 43 multi-image items from the 900 validation questions and report results on the remaining 857 items.
For fairness, all methods are evaluated on this filtered subset.
Following the official setup,\footnote{\url{https://github.com/MMMU-Benchmark/MMMU}} we append instruction postfixes: for open-ended questions we add ``\texttt{Answer the question using a single word or phrase.}''; for MC questions we add ``\texttt{Answer with the option's letter from the given choices directly.}'' to adopt official parsing tool. 

\textit{Metrics.}
Let $\mathcal{S}$ be the set of 30 subjects and let $\mathcal{D}_s$ denote the items belonging to subject $s \in \mathcal{S}$ after filtering.
For each subject $s$, we compute the subject-wise accuracy
\begin{equation*}
    \text{Acc}_{\mathcal{M}}(s)
    \;=\;
    \frac{1}{|\mathcal{D}_s|} \sum_{x \in \mathcal{D}_s}
    \mathbf{1}\!\left\{\,\mathcal{M}(x) = y(x)\,\right\},
\end{equation*}
where $y(x)$ is the ground-truth answer and $\mathcal{M}(x)$ is the model prediction.
We then report the macro-averaged accuracy over subjects,
\begin{equation*}
    \text{Accuracy}
    \;=\;
    \frac{1}{|\mathcal{S}|}
    \sum_{s \in \mathcal{S}}
    \text{Acc}_{\mathcal{M}}(s),
\end{equation*}
which follows the official subject-level evaluation protocol.

\paragraph{LLaVA-Bench-In-the-Wild dataset.}
We additionally employ the LLaVA-Bench-In-the-Wild~\citep{liu2023visual} dataset to qualitatively evaluate the visual grounding ability of LVLMs in open-ended image understanding. 
The dataset consists of 24 diverse images, including both real-world photographs and abstract illustrations, and 60 textual prompts designed to elicit complex visual reasoning and language generation.
Following prior work~\citep{leng2024vcd}, we present representative qualitative examples to illustrate how different decoding strategies affect grounding fidelity and response quality in challenging scenarios.

\subsection{Implementation details}
\label{appendix:implementation-details}
\paragraph{Model setup.}
All LVLMs are used with their official checkpoints and default inference configurations, including precision (e.g., \texttt{fp16}/\texttt{bf16}), attention mechanism, and tokenizer settings, unless otherwise specified.

\paragraph{Baseline decoding methods.}
For all decoding baselines, we primarily adopt the official inference code and default hyperparameters released by the authors.
When an official implementation for a given LVLM backbone is unavailable or cannot be directly applied, we instead follow the configuration described in the original paper.
In particular, for SID~\citep{huo2024sid} we follow the official LLaVA implementation, which uses a fixed budget of visual tokens, and for LVLMs whose number of visual tokens varies with the input resolution we select approximately $10\%$ of the visual tokens per image, consistent with the paper description.

\paragraph{\name{} hyperparameters.}
\name{} involves two hyperparameters: (1) the pool of decoder layer(s) from which vision tokens are selected and (2) the alignment weight $\alpha$ that determines the size of the constrained vocabulary subset.
For LLaVA-1.5-7B and InternVL3-8B, we select vision tokens from the final decoder layer (\texttt{last}) and fix this choice across all benchmarks.
For Qwen2.5-VL-7B, aggregating vision tokens from all even-numbered decoder layers (\texttt{all}) consistently performs better, and we use this configuration for all tasks.
The threshold $\alpha$ controls the tightness of the constrained vocabulary subset $\mathcal{V}_{\texttt{cons}}^t$: smaller $\alpha$ produces a looser subset with higher candidate diversity, whereas larger $\alpha$ produces a tighter subset that suppresses distractors more aggressively. Empirically, generative captioning tends to benefit from relatively looser subsets to preserve grounded coverage, while short-answer or discriminative settings often benefit from tighter subsets to improve precision. We select $\alpha$ from a logarithmic grid between $10^{-1}$ and $10^{-6}$ on a small validation split; when task-specific tuning is not required, we use $\alpha = 10^{-5}$ as the default.

\clearpage
\onecolumn
\section{Detailed Quantitative Results}
\label{appendix:quanti}

\subsection{Comprehensive results across eight models}
\input{Tables/Models}

\clearpage
\subsection{Comprehensive POPE results}
\label{appendix:pope}
\input{Tables/POPE}

\clearpage
\subsection{Comprehensive MMMU results}
\input{Tables/MMMU}

\clearpage
\twocolumn
\subsection{Token-level computation speed}
\label{appendix:computation}
All measurements are conducted on an NVIDIA RTX A6000 GPU with 48GB memory and an Intel(R) Xeon(R) Gold 5320 CPU running at 2.20 GHz.
We perform 6 warmup iterations before measurement to eliminate initial overhead, and each of the 300 evaluation samples is decoded three times per method.
We report the average decoding time per generated token, measured in milliseconds (ms/token).
Other evaluation configurations, including preprocessing pipelines and evaluation scripts, follow the official CHAIR benchmark implementation.
\input{Tables/computation}

\subsection{Comprehensive ablation results}
\label{appendix:ablation}
This section provides comprehensive ablation results complementing Sec.~\ref{sec:experiment_additional}.

We first report CHAIR ablations on vocabulary constraint, token selection, layer scope, and threshold $\alpha$, together with additional design-choice comparisons on selection metrics and fusion strategies (Table~\ref{tab:ablation}).

For selection, we compare alternative distance measures, including KL divergence and cosine similarity, to test whether the effectiveness of the token-selection step depends strongly on the exact metric choice or whether JSD mainly serves as a stable default.

For refinement, let \(p_{\text{output}}\) denote the vanilla next-token distribution on \(\mathcal{V}^t_{\texttt{cons}}\), and let \(p_{\text{vision}}\) denote the selected vision-token distribution on the same constrained set. We then compare interpolation and Product-of-Experts (PoE) fusion under different coefficients. Specifically, we consider
\[
    (1-\lambda)\,p_{\text{output}} + \lambda\,p_{\text{vision}}
\]
for interpolation, and
\[
    \log p_{\text{output}} + \lambda \log p_{\text{vision}}
\]
for PoE fusion.
This PoE-style form is also consistent with the KL-regularized interpretation of the refinement rule in Eq.~\eqref{eq:log-space-addition}; a detailed derivation is provided in Appendix~\ref{appendix:revisit_theory}.
In interpolation, $\lambda=0.5$ corresponds to balanced mixing, while $\lambda=0.25$ and $\lambda=0.75$ place relatively more weight on the output and vision distributions, respectively.
In PoE fusion, $\lambda=1.0$ corresponds to the default equal-strength combination, while smaller values weaken and larger values strengthen the influence of the vision-induced distribution.

As shown in Table~\ref{tab:ablation}, JSD provides the strongest overall trade-off among the tested selection metrics, while KL divergence and cosine similarity remain competitive but consistently yield weaker hallucination reduction and lower object-level F1. For refinement, the default PoE fusion performs best overall; interpolation remains competitive near balanced mixing, but is generally weaker, and both overly weak and overly strong PoE coefficients degrade performance.

We then present additional VQAv2 ablation results on layer-scope and $\alpha$ choices to show that the same trend persists on a discriminative benchmark (Table~\ref{tab:ablation_vqa}).
\input{Tables/Ablation}

\clearpage
\section{Detailed Qualitative Results}

\input{Tables/appendix_fig1}
\input{Tables/appendix_fig2_1}
\input{Tables/appendix_fig2_2}
\input{Figures/quali_noAPC}
\input{Figures/quali_ex1}
\input{Figures/quali_ex2}

\subsection{Vision token analysis case studies}
\label{appendix:motivation-fig}
Building on the decoding procedure in Appendix~\ref{appendix:revisit}, 
we now examine two representative case studies that instantiate our vision token analysis:
the fruit-counting example in Fig.~\ref{fig:overview} and the landscape description in Fig.~\ref{fig:motivation}.
For both cases, we report token-level distributions and positional metadata to make the contribution of individual vision tokens explicit.

All examples are drawn from the LLaVA-Bench-In-the-Wild dataset~\citep{liu2023visual} using LLaVA-1.5-7B. 
Vision token embeddings are extracted from the final decoder layer ($L = 32$), and the threshold for constrained vocabulary selection is set to $\alpha = 10^{-5}$, consistent with our default generative configuration.

\paragraph{Fruit-counting example (Fig.~\ref{fig:overview}).}
This example corresponds to the decoding step $t = 2$ described in Appendix~\ref{appendix:revisit}.
At this step, \name{} selects vision token index \texttt{229} as the most relevant reference based on the context-aware constrained distribution over $\mathcal{V}_{\texttt{cons}}^2$.
Table~\ref{tab:appendix_fig1} reports the log-probability values from three sources: 
(1) the base model output (Eq.~\eqref{eq:softmax}), 
(2) the vision-token projection at index \texttt{229} (Eq.~\eqref{eq:vision_project}), and 
(3) the final combined distribution after fusion (Eq.~\eqref{eq:final_fusion}).
Within the constrained subset, the vision token sharpens the distribution over numerals: it boosts the probability of ``\texttt{three}'' while suppressing ``\texttt{four}'' and other alternatives, thereby aligning the next-token prediction with the visual evidence.

\paragraph{Landscape description example (Fig.~\ref{fig:motivation}).}
Figure~\ref{fig:motivation} visualizes a complementary case from the same dataset (image ID: \texttt{003.jpg}, prompt: ``\texttt{Describe this photo in detail.}’’). 
Here we focus on how the semantics of individual vision tokens become more interpretable under vocabulary constraints.

The three tokens shown on the \textbf{left} of the figure (pink, yellow, cyan) are used to examine the semantic alignment of vision-token projections.
For each token, we compute the projected distribution over (i) the full vocabulary $\mathcal{V}$ and (ii) a manually defined constrained subset 
$\mathcal{V}_{\texttt{env}} = \{\texttt{river, mountain, person, sky, tree,}$ $\texttt{cloud, sea, grass, rock}\}$.
Table~\ref{tab:appendix_fig2_1} lists the corresponding log-probabilities and probabilities.
When restricted to $\mathcal{V}_{\texttt{env}}$, each vision token assigns most of its mass to a small number of semantically coherent labels (e.g., ``\texttt{sky}'' for the pink token, ``\texttt{river}''/``\texttt{rock}'' for the yellow token, ``\texttt{person}'' for the cyan token), 
making their textual semantics explicit and interpretable.

The token on the \textbf{right} (vision token index \texttt{193}) demonstrates how \name{} uses a single vision token as a reference signal during decoding.
At generation step $t = 72$, this token acts as the selected reference when refining the output distribution over $\mathcal{V}_{\texttt{cons}}^{72}$.
Table~\ref{tab:appendix_fig2_2} provides the detailed scores.
The probability of ``\texttt{painting}'' increases substantially after fusion, while the hallucinatory alternative ``\texttt{person}'' is suppressed, showing how context-aware constrained divergence encourages visually grounded descriptions without additional training.

\subsection{Case study without vocabulary subset}
\label{appendix:wo-vocab}

This case study illustrates how crucial the vocabulary subset constraint is for referencing vision tokens in logit lens manner. Figure~\ref{fig:noAPC} shows three qualitative comparison between with and without vocabulary subset.

When the subset is removed (\emph{w/o subset}), the model quickly diverges into degenerate behavior:
it produces partially fluent prefixes but then collapses into long, repetitive sequences of a single token or meaningless character streams until the maximum length is reached.
In contrast, with the subset enabled (\emph{w/ subset}), the model generates coherent, visually grounded descriptions that correctly capture the scene content.

These examples highlight a failure mode of directly projecting vision tokens onto the full vocabulary space.
Without the subset, the projected logits can over-amplify a few high-frequency or locally consistent tokens that are unrelated to the image, causing the fused distribution to collapse and reinforcing the same token at every subsequent step.
By constraining the output space to a contextually plausible vocabulary subset, \name{} prevents this collapse, keeps probability mass concentrated on semantically relevant candidates, and preserves both fluency and visual faithfulness during generation.

\subsection{Additional qualitative examples}
\label{appendix:quali}
To further illustrate the qualitative behavior of \name{}, we present two representative examples, including a sample from the LLaVA-Bench-In-The-Wild dataset, which contains diverse real-world images requiring complex reasoning beyond conventional benchmarks.

In the first example (Fig.~\ref{fig:quali_ex1}), the input is a cartoon-style illustration that contrasts classical statistical learning with neural networks via a visual metaphor. 
Baseline methods fail to capture this high-level analogy: they either focus on local, literal details of the scene or fall into repetitive, generic phrasing, thereby missing the notion of ``stacking more layers.'' 
In contrast, \name{} correctly identifies that the comic is about neural network layers and produces a coherent description that explicitly aligns the visual elements (e.g., the whiteboard and stacked layers) with the intended concept.

The second example (Fig.~\ref{fig:quali_ex2}) is an illustration of a bear, a cat, and a rabbit seated around a table with a plate of donuts. 
Here, baselines introduce object- and attribute-level hallucinations, such as mentioning cookies or pink cups that are not present in the image. 
\name{}, on the other hand, provides a faithful caption that correctly enumerates the animals and their activities without introducing unsupported details, demonstrating improved visual grounding in everyday scenes as well.

\newpage
\section{Usage of AI assistants}
In preparing this work, we utilized AI-based writing assistants in a limited way to suggest alternative phrasings, correct grammatical errors, and improve the readability.
All research ideas, experimental designs, implementation details, and reported results were determined, implemented, and verified by the authors, and all model outputs and quantitative results in the paper were generated and checked through our own code and experiments.
AI assistants did not generate any factual content reported in the paper (e.g., experimental results, dataset statistics, or citations), ensuring the originality, scientific contributions, technical content, methodology, and experimental findings are entirely attributable to the authors.

%% file: Tex/Algorithm.tex
\begin{algorithm*}[t]
\caption{\name{} decoding algorithm}
\label{alg:main}
\begin{algorithmic}
    \State \textbf{Input:} LVLM $\mathcal{M}$ with final decoder layer $L$, projection (language modeling) head $W_{\texttt{proj}}$, and vocabulary $\mathcal{V}$), input image $I$, set of candidate decoder layers $\mathcal{J}$, input text prompt $x_{\text{raw}}$
    \vspace{0.05in}
    \hrule
    \vspace{0.05in}
    \State Tokenize input text: $\mathbf{x} = \{x_0, \dots, x_{|\mathbf{x}| - 1}\} \leftarrow \texttt{Tokenizer}(x_{\text{raw}})$
    \State Encode image: $\mathbf{v} = \{v_0, \dots, v_{|\mathbf{v}| - 1}\} \leftarrow \texttt{VisionEncoder}(I)$
    \State Construct input sequence: $\mathbf{z} = \mathbf{v} \mathbin\Vert \mathbf{x}$ \Comment{$\mathbf{z} = \{z_0, \dots, z_{T-1}\}$ where $T = |\mathbf{v}| + |\mathbf{x}|$}
    \State Initialize output sequence $\mathbf{y} \leftarrow \emptyset$ \Comment{$t \leftarrow 0$}
    \While{not end-of-sequence}
        \State Get decoder hidden states $h_{T+t-1}^L$ from $\mathcal{M}$ with context $(\mathbf{z}, \mathbf{y})$
        
        \If{first-timestep}
            \State Given $\{h_i^j\}$ for $i = 0, \dots, T - 1$ and $j \in \mathcal{J}$ from initial forward pass
            \State Initialize cache $\mathbf{l}_{\texttt{vision}} \leftarrow \emptyset$
            \For{each layer $j \in \mathcal{J}$}
                \For{$i = 0$ to $|\mathbf{v}| - 1$}
                    \State $l_i^j = W_{\texttt{proj}}^\top h_i^j$
                    \State Append $l_i^j$ to $\mathbf{l}_{\texttt{vision}}$
                \EndFor
            \EndFor
            \State Cache matrix $\mathbf{l}_{\texttt{vision}} \in \mathbb{R}^{|\mathcal{J}| \cdot |\mathbf{v}| \times |\mathcal{V}|}$
            \State first-timestep $\leftarrow$ False
        \EndIf
        
        \State Compute vanilla distribution $p_{\text{base}} = \texttt{softmax}_{\mathcal{V}}(W_{\texttt{proj}}^\top h_{T+t-1}^L)$ (Eq.~\eqref{eq:softmax})
        \State Select constrained vocabulary $\mathcal{V}_{\texttt{cons}}^t \subseteq \mathcal{V}$ based on $p_{\text{base}}$ (Eq.~\eqref{eq:our_apc})
        \State Let $p_{\text{base}}^{\texttt{cons}} = \texttt{softmax}_{\mathcal{V}_{\texttt{cons}}^t}(W_{\texttt{proj}}^\top h_{T+t-1}^L)$

        \State $(j^*, i^*) \leftarrow \arg\min_{(j, i)} \texttt{JSD}(p_{\text{base}}^{\texttt{cons}} \Vert \texttt{softmax}_{\mathcal{V}_{\texttt{cons}}^t}(\mathbf{l}_{\texttt{vision}}[j, i]))$

        \State Let $p_{\text{ref}}^{\texttt{cons}} = \texttt{softmax}_{\mathcal{V}_{\texttt{cons}}^t}(\mathbf{l}_{\texttt{vision}}[j^*, i^*])$
        \State 
        Logit adjustment via log-space addition: $l_{\text{ReVisiT}} = \log(p_{\text{base}}^{\texttt{cons}}) + \log(p_{\text{ref}}^{\texttt{cons}})$
        \State Normalize over constrained vocabulary: $p_{\text{ReVisiT}} = \texttt{softmax}_{\mathcal{V}_{\texttt{cons}}^t}(l_{\text{ReVisiT}})$
        \State Sample next token: $y_t \sim p_{\text{ReVisiT}}$
        \State Append $y_t$ to $\mathbf{y}$
        \State $t \leftarrow t + 1$
    \EndWhile
    \State \textbf{return} $\mathbf{y}$
\end{algorithmic}
\end{algorithm*}
\vspace{-100cm}

%% file: Tables/Models.tex
\begin{table*}[h]
\begin{center}
\begin{small}

\setlength{\tabcolsep}{10pt}
\caption{\textbf{Comprehensive CHAIR results across various model sizes and architectures.}
Lower scores ($\downarrow$) on CHAIR$_S$, CHAIR$_I$ and higher ($\uparrow$) F1 indicate better performance. The best results in each model are \textbf{bolded}.}
\vspace{0.1in}
\label{tab:appendix_chair_by_model}
\begin{tabular}{l l ccc}
    \toprule
    \textbf{Model} & \textbf{Method} & CHAIR$_S$ $\downarrow$ & CHAIR$_I$ $\downarrow$ & F1 $\uparrow$ \\
    \midrule
    LLaVA-1.5-7B
    & Greedy                  & \tc{53.8} & \tc{14.66} & \tc{82.33} \\
    \citep{liu2024improved} & \ca{Ours} & \ca{50.6} & \ca{13.43} & \ca{83.17} \\
    \midrule
    Qwen2.5-VL-7B
    & Greedy                  & \tc{35.2} & \tc{8.43}  & \tc{79.85} \\
    \citep{bai2025qwen2} & \ca{Ours} & \ca{29.8} & \ca{7.04}  & \ca{81.16} \\
    \midrule
    InternVL2.5-8B
    & Greedy                  & \tc{33.4} & \tc{8.47}  & \tc{80.58} \\
    \citep{chen2024internvl25} & \ca{Ours} & \ca{29.8} & \ca{7.81}  & \ca{81.93} \\
    \midrule
    InternVL3-8B
    & Greedy                  & \tc{29.4} & \tc{7.91}  & \tc{81.11} \\
    \citep{zhu2025internvl3} & \ca{Ours} & \ca{27.6} & \ca{7.27}  & \ca{82.29} \\
    \midrule
    LLaVA-1.5-13B
    & Greedy                  & \ta{48.0} & \tc{13.29} & \tc{83.27} \\
    \citep{liu2024improved} & \ca{Ours} & \cc{52.6} & \ca{13.25} & \ca{83.50} \\
    \midrule
    InternVL3-14B
    & Greedy                  & \tc{33.2} & \tc{9.34}  & \tc{81.49} \\
    \citep{zhu2025internvl3} & \ca{Ours} & \ca{33.0} & \ca{8.66}  & \ca{82.03} \\
    \midrule
    InternVL2.5-26B
    & Greedy                  & \tc{31.8} & \tc{8.39}  & \tc{79.60} \\
    \citep{chen2024internvl25} & \ca{Ours} & \ca{30.4} & \ca{7.42}  & \ca{80.71} \\
    \midrule
    Qwen2.5-VL-32B
    & Greedy                  & \tc{53.6} & \tc{8.85}  & \tc{82.00} \\
    \citep{bai2025qwen2} & \ca{Ours} & \ca{53.4} & \ca{8.5}   & \ca{82.55} \\
    \bottomrule
\end{tabular}
\end{small}
\end{center}
\vspace{-8pt}
\end{table*}

%% file: Tables/POPE.tex
\begin{table*}[h]
\begin{center}
\begin{small}
\setlength{\tabcolsep}{8pt}
\caption{
    \textbf{Comprehensive results on POPE benchmark} across datasets (MS-COCO, A-OKVQA, GQA) and setups (Random / Popular / Adversarial).
        Higher scores ($\uparrow$) on Accuracy, Precision, and F1 indicate better performance. 
}
\label{tab:POPE_full}
\resizebox{0.9\textwidth}{!}{
\begin{tabular}{lllccccccccc}
\toprule
\multirow{2}{*}[-2pt]{\textbf{Dataset}} 
& \multirow{2}{*}[-2pt]{\textbf{Setup}} 
& \multirow{2}{*}[-2pt]{\textbf{Method}} 
& \multicolumn{3}{c}{\textbf{LLaVA-1.5-7B}} 
& \multicolumn{3}{c}{\textbf{Qwen2.5-VL-7B}} 
& \multicolumn{3}{c}{\textbf{InternVL3-8B}} \\
\cmidrule(lr){4-6}\cmidrule(lr){7-9}\cmidrule(lr){10-12}
& & & Acc $\uparrow$ & Prec $\uparrow$ & F1 $\uparrow$
& Acc $\uparrow$ & Prec $\uparrow$ & F1 $\uparrow$
  & Acc $\uparrow$ & Prec $\uparrow$ & F1 $\uparrow$ \\
\midrule

\multirow{21}{*}{MS-COCO} & \multirow{7}{*}{Random} & Greedy & \tc{89.07} & \tc{89.54} & \tb{89.00} & \tc{84.07} & \tb{99.71} & \tc{81.09} & \tb{94.53} & \tb{97.31} & \tb{94.37} \\
& & DoLa & \tc{89.00} & \tc{89.58} & \tc{88.92} & \tc{81.53} & \ta{99.89} & \tc{77.37} & \tc{91.17} & \ta{98.43} & \tc{90.45} \\
& & VCD & \tc{87.23} & \tc{86.62} & \tc{87.34} & \tb{84.57} & \tc{99.15} & \tb{81.88} & \tc{93.00} & \tc{95.62} & \tc{92.79} \\
& & M3ID & \tc{89.07} & \tc{89.54} & \tb{89.00} & \tc{83.93} & \tc{99.32} & \tc{80.96} & \tc{94.33} & \tc{97.30} & \tc{94.15} \\
& & CODE & \tc{86.53} & \tc{87.23} & \tc{86.41} & \tc{84.00} & \tc{98.02} & \tc{81.26} & \tc{93.70} & \tc{94.50} & \tc{93.64} \\
& & SID & \ta{89.67} & \ta{91.49} & \ta{89.43} & \ta{86.93} & \tc{99.03} & \ta{85.10} & \tc{92.97} & \tc{95.48} & \tc{92.77} \\
& & \ca{Ours} & \cb{89.10} & \cb{89.82} & \cb{89.00} & \cc{84.30} & \cb{99.71} & \cc{81.42} & \ca{94.57} & \cc{97.18} & \ca{94.41} \\
\cmidrule(lr){2-12}
& \multirow{7}{*}{Popular} & Greedy & \tc{85.63} & \tc{83.72} & \tc{86.03} & \tc{83.47} & \tc{97.99} & \tc{80.52} & \tb{90.57} & \tc{89.75} & \tb{90.66} \\
& & DoLa & \tc{85.63} & \tc{83.85} & \tc{86.00} & \tc{81.10} & \ta{98.54} & \tc{76.96} & \tc{89.20} & \ta{94.08} & \tc{88.57} \\
& & VCD & \tc{83.53} & \tc{80.75} & \tc{84.25} & \tb{83.93} & \tc{97.39} & \tb{81.27} & \tc{88.57} & \tc{87.39} & \tc{88.74} \\
& & M3ID & \tc{85.63} & \tc{83.72} & \tc{86.03} & \tc{83.40} & \tc{97.81} & \tc{80.46} & \tc{90.43} & \tc{89.82} & \tc{90.51} \\
& & CODE & \tc{84.73} & \tc{84.14} & \tc{84.86} & \tc{83.63} & \tc{97.02} & \tc{80.92} & \tc{88.47} & \tc{85.40} & \tc{88.95} \\
& & SID & \ta{86.13} & \ta{85.19} & \ta{86.32} & \ta{86.23} & \tc{97.22} & \ta{84.42} & \tc{88.53} & \tc{87.29} & \tc{88.72} \\
& & \ca{Ours} & \cb{85.80} & \cb{84.16} & \cb{86.13} & \cc{83.70} & \cb{98.01} & \cc{80.85} & \ca{90.67} & \cb{89.87} & \ca{90.76} \\
\cmidrule(lr){2-12}
& \multirow{7}{*}{Adversarial} & Greedy & \tc{79.27} & \tc{74.72} & \tc{81.01} & \tc{83.10} & \tc{97.06} & \tc{80.16} & \tc{87.10} & \tc{84.16} & \tb{87.63} \\
& & DoLa & \tc{79.40} & \tc{74.97} & \tc{81.08} & \tc{80.83} & \ta{97.73} & \tc{76.71} & \ta{87.30} & \ta{90.34} & \tc{86.80} \\
& & VCD & \tc{77.67} & \tc{72.90} & \tc{79.77} & \tb{83.37} & \tc{95.96} & \tb{80.73} & \tc{86.13} & \tc{83.33} & \tc{86.69} \\
& & M3ID & \tc{79.27} & \tc{74.72} & \tc{81.01} & \tc{83.00} & \tc{96.79} & \tc{80.06} & \tc{87.03} & \tb{84.31} & \tc{87.53} \\
& & CODE & \tc{78.67} & \tc{75.23} & \tc{80.02} & \tc{83.00} & \tc{95.41} & \tc{80.31} & \tc{85.00} & \tc{80.24} & \tc{86.09} \\
& & SID & \ta{80.30} & \ta{76.56} & \ta{81.59} & \ta{85.17} & \tc{94.74} & \ta{83.39} & \tc{85.63} & \tc{82.61} & \tc{86.27} \\
& & \ca{Ours} & \cb{79.90} & \cb{76.21} & \cb{81.22} & \cb{83.37} & \cb{97.08} & \cc{80.53} & \cb{87.13} & \cc{84.05} & \ca{87.69} \\
\midrule
\multirow{21}{*}{A-OKVQA} & \multirow{7}{*}{Random} & Greedy & \tc{86.33} & \tc{81.00} & \tc{87.42} & \tc{87.80} & \tc{96.86} & \tc{86.49} & \tb{94.00} & \tb{93.25} & \tb{94.05} \\
& & DoLa & \tc{86.43} & \tc{81.25} & \tc{87.47} & \tc{85.33} & \ta{96.99} & \tc{83.26} & \tc{91.10} & \ta{95.50} & \tc{90.65} \\
& & VCD & \tc{83.70} & \tc{77.91} & \tc{85.23} & \tc{87.87} & \tc{96.56} & \tc{86.62} & \tc{90.17} & \tc{88.65} & \tc{90.36} \\
& & M3ID & \tc{86.33} & \tc{81.00} & \tc{87.42} & \tc{87.40} & \tc{95.91} & \tc{86.11} & \tb{94.00} & \tb{93.25} & \tb{94.05} \\
& & CODE & \tc{83.80} & \tc{78.48} & \tc{85.18} & \tc{87.50} & \tc{93.64} & \tc{86.55} & \tc{91.20} & \tc{87.50} & \tc{91.61} \\
& & SID & \ta{87.57} & \tb{83.48} & \ta{88.28} & \ta{88.83} & \tc{96.12} & \ta{87.88} & \tc{89.70} & \tc{87.81} & \tc{89.95} \\
& & \ca{Ours} & \cb{87.37} & \ca{84.75} & \cb{87.83} & \cb{88.27} & \cb{96.90} & \cb{87.08} & \ca{94.17} & \cc{93.16} & \ca{94.23} \\
\cmidrule(lr){2-12}
& \multirow{7}{*}{Popular} & Greedy & \tc{78.77} & \tc{71.74} & \tc{81.72} & \tc{85.93} & \ta{92.58} & \tc{84.74} & \tb{89.67} & \tc{85.93} & \tc{90.18} \\
& & DoLa & \tc{79.00} & \tc{72.06} & \tc{81.85} & \tc{83.37} & \tc{92.17} & \tc{81.43} & \tc{89.23} & \ta{91.71} & \tc{88.90} \\
& & VCD & \tc{76.47} & \tc{69.58} & \tc{79.99} & \tc{85.90} & \tc{92.10} & \tc{84.78} & \tc{81.90} & \tc{76.48} & \tc{83.58} \\
& & M3ID & \tc{78.77} & \tc{71.74} & \tc{81.72} & \tc{85.77} & \tc{92.21} & \tc{84.59} & \ta{89.70} & \tb{85.98} & \tb{90.21} \\
& & CODE & \tc{77.70} & \tc{71.17} & \tc{80.68} & \tc{86.07} & \tc{90.62} & \tb{85.24} & \tc{83.57} & \tc{76.82} & \tc{85.40} \\
& & SID & \tb{80.63} & \tb{74.30} & \tb{82.87} & \ta{87.00} & \tc{92.11} & \ta{86.16} & \tc{81.60} & \tc{76.07} & \tc{83.36} \\
& & \ca{Ours} & \ca{81.97} & \ca{77.54} & \ca{83.31} & \cb{86.30} & \cb{92.44} & \cc{85.23} & \cb{89.67} & \cc{85.63} & \ca{90.22} \\
\cmidrule(lr){2-12}
& \multirow{7}{*}{Adversarial} & Greedy & \tc{68.20} & \tc{61.86} & \tc{74.91} & \tc{80.83} & \tc{82.59} & \tc{80.30} & \tc{82.83} & \tc{76.46} & \tc{84.68} \\
& & DoLa & \tc{68.33} & \tc{62.00} & \tc{74.95} & \tc{79.63} & \ta{84.22} & \tc{78.17} & \ta{84.43} & \ta{83.22} & \tc{84.71} \\
& & VCD & \tc{67.80} & \tc{61.69} & \tc{74.47} & \tc{81.10} & \tc{82.69} & \tc{80.63} & \tc{78.07} & \tc{71.84} & \tc{80.81} \\
& & M3ID & \tc{68.20} & \tc{61.86} & \tc{74.91} & \tc{79.90} & \tc{81.00} & \tc{79.54} & \tc{82.97} & \tc{76.63} & \tb{84.78} \\
& & CODE & \tc{68.30} & \tc{62.23} & \tc{74.61} & \tc{79.37} & \tc{78.73} & \tc{79.59} & \tc{79.00} & \tc{71.60} & \tc{82.07} \\
& & SID & \tb{71.33} & \tb{64.75} & \tb{76.57} & \ta{82.40} & \tb{83.38} & \ta{82.14} & \tc{77.43} & \tc{71.18} & \tc{80.34} \\
& & \ca{Ours} & \ca{72.83} & \ca{67.00} & \ca{76.81} & \cb{81.33} & \cc{82.82} & \cb{80.90} & \cb{83.07} & \cb{76.70} & \ca{84.87} \\
\midrule
\multirow{21}{*}{GQA} & \multirow{7}{*}{Random} & Greedy & \tc{85.97} & \tc{79.72} & \tc{87.30} & \tc{87.53} & \ta{97.15} & \tc{86.12} & \tb{90.47} & \tc{95.98} & \tb{89.86} \\
& & DoLa & \tb{86.07} & \tb{79.92} & \tb{87.36} & \tc{83.47} & \tc{94.66} & \tc{81.10} & \tc{84.03} & \tc{95.70} & \tc{81.70} \\
& & VCD & \tc{83.77} & \tc{77.16} & \tc{85.53} & \tb{87.83} & \tc{96.78} & \tb{86.55} & \tc{85.50} & \tc{95.55} & \tc{83.70} \\
& & M3ID & \tc{86.00} & \tc{79.74} & \tc{87.33} & \tc{87.43} & \tc{96.14} & \tc{86.12} & \tc{90.20} & \tc{95.89} & \tc{89.55} \\
& & CODE & \tc{84.87} & \tc{78.58} & \tc{86.37} & \tc{87.07} & \tc{94.48} & \tc{85.89} & \tc{88.80} & \tc{95.68} & \tc{87.89} \\
& & SID & \tc{85.97} & \tc{79.72} & \tc{87.30} & \ta{88.80} & \tc{96.78} & \ta{87.76} & \tc{86.07} & \ta{96.96} & \tc{84.24} \\
& & \ca{Ours} & \ca{87.13} & \ca{83.47} & \ca{87.80} & \cc{87.80} & \cb{97.09} & \cc{86.46} & \ca{90.80} & \cb{96.08} & \ca{90.24} \\
\cmidrule(lr){2-12}
& \multirow{7}{*}{Popular} & Greedy & \tc{73.80} & \tc{66.38} & \tc{78.64} & \tb{83.80} & \ta{88.82} & \tc{82.68} & \ta{88.27} & \tc{91.41} & \ta{87.80} \\
& & DoLa & \tb{73.97} & \tc{66.56} & \tb{78.73} & \tc{78.57} & \tc{83.71} & \tc{76.80} & \tc{82.57} & \tb{92.08} & \tc{80.35} \\
& & VCD & \tc{71.87} & \tc{64.76} & \tc{77.32} & \tc{83.33} & \tc{87.09} & \tc{82.44} & \tc{83.80} & \tc{91.56} & \tc{82.13} \\
& & M3ID & \tc{73.83} & \tc{66.39} & \tc{78.67} & \tc{80.67} & \tc{82.39} & \tc{80.14} & \tc{87.83} & \tc{90.97} & \tc{87.35} \\
& & CODE & \tc{73.90} & \tb{66.60} & \tc{78.60} & \tc{81.47} & \tc{83.29} & \tc{80.95} & \tc{86.60} & \tc{90.97} & \tc{85.85} \\
& & SID & \tc{73.80} & \tc{66.38} & \tc{78.64} & \ta{84.27} & \tc{87.25} & \ta{83.61} & \tc{84.40} & \ta{92.93} & \tc{82.68} \\
& & \ca{Ours} & \ca{78.67} & \ca{72.42} & \ca{81.28} & \cc{83.77} & \cb{88.28} & \cb{82.75} & \cb{87.93} & \cc{90.24} & \cb{87.58} \\
\cmidrule(lr){2-12}
& \multirow{7}{*}{Adversarial} & Greedy & \tc{68.17} & \tc{61.60} & \tc{75.19} & \tc{81.50} & \tc{84.36} & \tc{80.70} & \ta{85.13} & \tc{85.61} & \ta{85.03} \\
& & DoLa & \tc{68.40} & \tc{61.80} & \tc{75.30} & \tc{78.37} & \tc{83.32} & \tc{76.63} & \tc{80.63} & \tb{87.69} & \tc{78.63} \\
& & VCD & \tc{67.63} & \tc{61.22} & \tc{74.82} & \tb{82.00} & \tb{84.63} & \tb{81.29} & \tc{81.33} & \tc{86.55} & \tc{79.90} \\
& & M3ID & \tc{68.20} & \tc{61.62} & \tc{75.22} & \tc{80.50} & \tc{82.11} & \tc{80.00} & \tb{85.00} & \tc{85.71} & \tc{84.85} \\
& & CODE & \tb{68.70} & \tb{62.12} & \tb{75.39} & \tc{79.87} & \tc{80.56} & \tc{79.64} & \tc{83.93} & \tc{85.85} & \tc{83.49} \\
& & SID & \tc{68.17} & \tc{61.60} & \tc{75.19} & \ta{82.37} & \tc{83.79} & \ta{81.99} & \tc{82.07} & \ta{87.81} & \tc{80.59} \\
& & \ca{Ours} & \ca{73.40} & \ca{66.91} & \ca{77.68} & \cc{81.93} & \ca{84.71} & \cc{81.18} & \cc{84.90} & \cc{84.78} & \cb{84.93} \\
\bottomrule
\end{tabular}
}
\end{small}
\end{center}
\vspace{-5in}
\end{table*}

%% file: Tables/MMMU.tex
\begin{table*}[h]
\begin{center}
\begin{small}
\setlength{\tabcolsep}{3pt}
\caption{
    \textbf{MMMU per-subject accuracy}. 
    We report \textit{Accuracy} (\%) for each subject, model, and decoding method.
    Higher scores ($\uparrow$) indicate better performance. 
}
\label{tab:mmmu_subject}
\resizebox{1.0\textwidth}{!}{
\begin{tabular}{l|ccccccc|ccccccc|ccccccc}
    \toprule
    \multirow{2}{*}{\textbf{Subject}}
        & \multicolumn{7}{c|}{\textbf{LLaVA-1.5-7B} (Accuracy $\uparrow$)} 
        & \multicolumn{7}{c|}{\textbf{Qwen2.5-VL-7B} (Accuracy $\uparrow$)} 
        & \multicolumn{7}{c}{\textbf{InternVL3-8B} (Accuracy $\uparrow$)} \\
        
        & Greedy & DoLa & VCD & M3ID & CODE & SID & \ca{Ours} 
        & Greedy & DoLa & VCD & M3ID & CODE & SID & \ca{Ours} 
        & Greedy & DoLa & VCD & M3ID & CODE & SID & \ca{Ours}  \\
    \midrule
        Accounting & \tb{16.67} & \tb{16.67} & \tb{16.67} & \tb{16.67} & \ta{20.00} & \tc{13.33} & \cb{16.67} & \ta{46.67} & \tc{36.67} & \tb{40.00} & \ta{46.67} & \tc{26.67} & \tc{36.67} & \cc{36.67} & \tc{26.67} & \tc{23.33} & \tc{26.67} & \tc{26.67} & \tc{26.67} & \ta{33.33} & \cb{30.00} \\
        Agriculture & \ta{41.38} & \ta{41.38} & \tb{37.93} & \ta{41.38} & \ta{41.38} & \tb{37.93} & \cb{37.93} & \ta{55.17} & \tc{48.28} & \ta{55.17} & \tb{51.72} & \tc{41.38} & \tc{44.83} & \ca{55.17} & \ta{55.17} & \tb{51.72} & \tb{51.72} & \ta{55.17} & \ta{55.17} & \ta{55.17} & \ca{55.17} \\
        Architecture and Engineering & \tc{31.03} & \tc{31.03} & \tc{34.48} & \tc{31.03} & \ta{41.38} & \tb{37.93} & \cc{31.03} & \ta{58.62} & \tb{51.72} & \ta{58.62} & \tb{51.72} & \tb{51.72} & \tc{44.83} & \cc{48.28} & \tb{37.93} & \tb{37.93} & \tb{37.93} & \tb{37.93} & \ta{41.38} & \tc{27.59} & \cc{34.48} \\
        Art & \ta{56.67} & \tb{53.33} & \tb{53.33} & \ta{56.67} & \tb{53.33} & \tc{50.00} & \cb{53.33} & \ta{73.33} & \ta{73.33} & \ta{73.33} & \ta{73.33} & \tc{60.00} & \ta{73.33} & \cb{70.00} & \ta{76.67} & \tc{70.00} & \ta{76.67} & \ta{76.67} & \tb{73.33} & \ta{76.67} & \cb{73.33} \\
        Art Theory & \ta{64.00} & \ta{64.00} & \tc{56.00} & \ta{64.00} & \tc{56.00} & \tb{60.00} & \ca{64.00} & \ta{88.00} & \ta{88.00} & \ta{88.00} & \tb{80.00} & \tc{76.00} & \ta{88.00} & \ca{88.00} & \tb{92.00} & \tb{92.00} & \tb{92.00} & \tb{92.00} & \ta{96.00} & \tb{92.00} & \cb{92.00} \\
        Basic Medical Science & \tc{43.33} & \tc{43.33} & \ta{53.33} & \tc{46.67} & \tc{46.67} & \tb{50.00} & \cc{46.67} & \tb{60.00} & \ta{63.33} & \tc{56.67} & \tc{53.33} & \tc{56.67} & \tc{56.67} & \cb{60.00} & \ta{66.67} & \ta{66.67} & \tb{60.00} & \ta{66.67} & \tb{60.00} & \tb{60.00} & \cb{60.00} \\
        Biology & \ta{17.86} & \ta{17.86} & \tb{14.29} & \ta{17.86} & \tc{10.71} & \tc{10.71} & \ca{17.86} & \ta{42.86} & \tb{39.29} & \ta{42.86} & \ta{42.86} & \tb{39.29} & \ta{42.86} & \ca{42.86} & \ta{53.57} & \ta{53.57} & \ta{53.57} & \ta{53.57} & \tb{46.43} & \ta{53.57} & \ca{53.57} \\
        Chemistry & \tb{20.00} & \tb{20.00} & \tb{20.00} & \tb{20.00} & \tb{20.00} & \tb{20.00} & \ca{24.00} & \tc{28.00} & \tc{28.00} & \tb{32.00} & \tc{28.00} & \tc{20.00} & \tb{32.00} & \ca{36.00} & \tc{28.00} & \tc{28.00} & \tc{28.00} & \tc{28.00} & \tc{24.00} & \ta{36.00} & \cb{32.00} \\
        Clinical Medicine & \tb{34.48} & \tb{34.48} & \tc{24.14} & \tb{34.48} & \ta{37.93} & \tc{27.59} & \cb{34.48} & \ta{62.07} & \tc{55.17} & \ta{62.07} & \tc{48.28} & \tc{51.72} & \tb{58.62} & \cb{58.62} & \tb{68.97} & \tb{68.97} & \ta{72.41} & \tb{68.97} & \tb{68.97} & \ta{72.41} & \cb{68.97} \\
        Computer Science & \ta{32.14} & \ta{32.14} & \tb{28.57} & \ta{32.14} & \tc{25.00} & \tb{28.57} & \ca{32.14} & \tb{50.00} & \ta{53.57} & \ta{53.57} & \tb{50.00} & \tc{39.29} & \tb{50.00} & \ca{53.57} & \tb{53.57} & \tc{50.00} & \tb{53.57} & \tb{53.57} & \tc{50.00} & \tc{50.00} & \ca{57.14} \\
        Design & \tb{50.00} & \tb{50.00} & \tb{50.00} & \tb{50.00} & \tb{50.00} & \ta{53.33} & \cc{46.67} & \ta{73.33} & \ta{73.33} & \ta{73.33} & \ta{73.33} & \tb{56.67} & \ta{73.33} & \ca{73.33} & \ta{73.33} & \ta{73.33} & \ta{73.33} & \ta{73.33} & \ta{73.33} & \ta{73.33} & \ca{73.33} \\
        Diagnostics and Laboratory Medicine & \ta{34.48} & \ta{34.48} & \ta{34.48} & \ta{34.48} & \tb{31.03} & \ta{34.48} & \ca{34.48} & \tb{41.38} & \tc{27.59} & \tb{41.38} & \tc{37.93} & \tc{34.48} & \tc{37.93} & \ca{44.83} & \tc{51.72} & \tc{51.72} & \tb{55.17} & \tc{51.72} & \tb{55.17} & \tc{44.83} & \ca{58.62} \\
        Economics & \ta{32.14} & \tb{28.57} & \tc{25.00} & \ta{32.14} & \ta{32.14} & \ta{32.14} & \ca{32.14} & \tb{42.86} & \tb{42.86} & \tc{35.71} & \tc{39.29} & \tc{28.57} & \ta{46.43} & \cb{42.86} & \ta{57.14} & \tb{53.57} & \ta{57.14} & \ta{57.14} & \ta{57.14} & \tb{53.57} & \ca{57.14} \\
        Electronics & \tb{26.67} & \tb{26.67} & \tb{26.67} & \tb{26.67} & \tb{26.67} & \tc{23.33} & \ca{30.00} & \tc{30.00} & \ta{36.67} & \ta{36.67} & \tc{30.00} & \tc{26.67} & \tb{33.33} & \ca{36.67} & \tb{26.67} & \ta{30.00} & \tc{20.00} & \tb{26.67} & \tc{20.00} & \tc{20.00} & \cb{26.67} \\
        Energy and Power & \tb{43.33} & \tb{43.33} & \ta{46.67} & \tb{43.33} & \ta{46.67} & \tb{43.33} & \ca{46.67} & \ta{33.33} & \ta{33.33} & \ta{33.33} & \ta{33.33} & \tb{30.00} & \tb{30.00} & \cb{30.00} & \tb{40.00} & \tb{40.00} & \tb{40.00} & \tb{40.00} & \tb{40.00} & \tc{36.67} & \ca{43.33} \\
        Finance & \ta{16.67} & \ta{16.67} & \ta{16.67} & \ta{16.67} & \ta{16.67} & \tb{13.33} & \ca{16.67} & \tc{23.33} & \ta{33.33} & \tb{26.67} & \tc{23.33} & \tc{23.33} & \tb{26.67} & \cb{26.67} & \tb{26.67} & \tc{23.33} & \tb{26.67} & \tc{23.33} & \tc{23.33} & \tc{23.33} & \ca{30.00} \\
        Geography & \tc{34.48} & \tc{34.48} & \tb{37.93} & \tc{34.48} & \ta{44.83} & \tc{34.48} & \cc{34.48} & \tb{37.93} & \tb{37.93} & \ta{41.38} & \ta{41.38} & \ta{41.38} & \tc{31.03} & \cb{37.93} & \ta{48.28} & \ta{48.28} & \ta{48.28} & \ta{48.28} & \ta{48.28} & \tb{37.93} & \ca{48.28} \\
        History & \tb{42.86} & \tb{42.86} & \tb{42.86} & \tb{42.86} & \tb{42.86} & \ta{46.43} & \ca{46.43} & \tb{67.86} & \tb{67.86} & \tc{64.29} & \tc{64.29} & \ta{75.00} & \tb{67.86} & \cb{67.86} & \ta{78.57} & \ta{78.57} & \tb{75.00} & \ta{78.57} & \tc{67.86} & \tb{75.00} & \cb{75.00} \\
        Literature & \tb{72.41} & \tb{72.41} & \ta{75.86} & \tb{72.41} & \tb{72.41} & \tb{72.41} & \cb{72.41} & \ta{82.76} & \ta{82.76} & \tb{79.31} & \tc{41.38} & \tc{44.83} & \tc{75.86} & \ca{82.76} & \tb{82.76} & \ta{86.21} & \tb{82.76} & \tb{82.76} & \tb{82.76} & \tb{82.76} & \ca{86.21} \\
        Manage & \tb{24.14} & \tb{24.14} & \ta{27.59} & \tb{24.14} & \tc{20.69} & \tc{20.69} & \cb{24.14} & \ta{44.83} & \tb{41.38} & \tb{41.38} & \tc{31.03} & \tc{37.93} & \tb{41.38} & \ca{44.83} & \tb{55.17} & \tb{55.17} & \tb{55.17} & \tb{55.17} & \tb{55.17} & \tc{51.72} & \ca{58.62} \\
        Marketing & \ta{26.67} & \ta{26.67} & \tc{20.00} & \ta{26.67} & \tb{23.33} & \ta{26.67} & \ca{26.67} & \tb{56.67} & \tc{53.33} & \ta{63.33} & \tb{56.67} & \tc{46.67} & \tb{56.67} & \ca{63.33} & \tb{56.67} & \ta{63.33} & \tb{56.67} & \tb{56.67} & \tb{56.67} & \tb{56.67} & \cc{53.33} \\
        Materials & \tb{25.93} & \ta{29.63} & \ta{29.63} & \ta{29.63} & \ta{29.63} & \ta{29.63} & \cb{25.93} & \tc{40.74} & \tc{29.63} & \ta{48.15} & \tc{40.74} & \tc{37.04} & \tb{44.44} & \cc{40.74} & \tb{40.74} & \tb{40.74} & \tb{40.74} & \tb{40.74} & \tb{40.74} & \tb{40.74} & \ca{44.44} \\
        Math & \ta{35.71} & \ta{35.71} & \tb{32.14} & \tb{32.14} & \tc{21.43} & \tc{28.57} & \ca{35.71} & \tc{42.86} & \tc{42.86} & \tc{32.14} & \tc{39.29} & \tb{46.43} & \ta{50.00} & \cb{46.43} & \ta{39.29} & \ta{39.29} & \tb{35.71} & \ta{39.29} & \ta{39.29} & \tb{35.71} & \ca{39.29} \\
        Mechanical Engineering & \ta{20.69} & \ta{20.69} & \ta{20.69} & \ta{20.69} & \ta{20.69} & \tb{13.79} & \ca{20.69} & \tc{37.93} & \tc{41.38} & \tc{37.93} & \tc{37.93} & \ta{48.28} & \tb{44.83} & \cc{41.38} & \tb{34.48} & \tb{34.48} & \tb{34.48} & \tb{34.48} & \tc{31.03} & \tb{34.48} & \ca{41.38} \\
        Music & \tc{36.00} & \tc{36.00} & \tc{36.00} & \tc{36.00} & \ta{44.00} & \tb{40.00} & \ca{44.00} & \tb{44.00} & \tc{40.00} & \tb{44.00} & \tb{44.00} & \tc{36.00} & \tc{36.00} & \ca{48.00} & \tc{40.00} & \tb{44.00} & \tc{36.00} & \tc{40.00} & \ta{48.00} & \tc{36.00} & \cc{36.00} \\
        Pharmacy & \tb{40.74} & \tb{40.74} & \tb{40.74} & \tb{40.74} & \ta{44.44} & \tc{37.04} & \cb{40.74} & \tc{55.56} & \ta{70.37} & \tb{62.96} & \tc{48.15} & \tc{44.44} & \tb{62.96} & \cc{55.56} & \tb{51.85} & \tb{51.85} & \tc{48.15} & \tb{51.85} & \tc{48.15} & \tb{51.85} & \ca{55.56} \\
        Physics & \tc{23.33} & \tc{23.33} & \tb{26.67} & \tc{23.33} & \tc{23.33} & \tc{23.33} & \ca{30.00} & \ta{36.67} & \ta{36.67} & \tb{33.33} & \tb{33.33} & \tb{33.33} & \tb{33.33} & \cc{30.00} & \ta{40.00} & \ta{40.00} & \ta{40.00} & \ta{40.00} & \ta{40.00} & \ta{40.00} & \ca{40.00} \\
        Psychology & \ta{34.62} & \ta{34.62} & \tc{26.92} & \ta{34.62} & \ta{34.62} & \tb{30.77} & \ca{34.62} & \ta{73.08} & \tb{69.23} & \ta{73.08} & \tc{57.69} & \tc{50.00} & \tb{69.23} & \ca{73.08} & \tb{76.92} & \tc{73.08} & \tb{76.92} & \tb{76.92} & \tb{76.92} & \ta{80.77} & \cb{76.92} \\
        Public Health & \tb{16.67} & \tb{16.67} & \tc{13.33} & \tb{16.67} & \ta{20.00} & \tb{16.67} & \cb{16.67} & \tc{50.00} & \ta{56.67} & \tb{53.33} & \tc{50.00} & \tc{46.67} & \tb{53.33} & \cc{50.00} & \ta{63.33} & \ta{63.33} & \ta{63.33} & \ta{63.33} & \ta{63.33} & \ta{63.33} & \ca{63.33} \\
        Sociology & \ta{36.67} & \ta{36.67} & \ta{36.67} & \ta{36.67} & \ta{36.67} & \ta{36.67} & \ca{36.67} & \ta{53.33} & \tb{50.00} & \tb{50.00} & \tb{50.00} & \tc{33.33} & \tb{50.00} & \cb{50.00} & \ta{63.33} & \tb{60.00} & \tb{60.00} & \ta{63.33} & \ta{63.33} & \tb{60.00} & \cb{60.00} \\
    \midrule
        Overall & \tc{34.39} & \tc{34.29} & \tc{33.51} & \tb{34.51} & \tc{34.48} & \tc{33.10} & \ca{35.13} & \tc{51.11} & \tc{50.15} & \tb{51.13} & \tc{46.63} & \tc{42.79} & \tc{49.75} & \ca{51.18} & \tb{53.54} & \tc{53.08} & \tc{52.60} & \tc{53.43} & \tc{52.41} & \tc{51.85} & \ca{54.14} \\
    \bottomrule
\end{tabular}
}
\end{small}
\end{center}
\end{table*}

%% file: Tables/computation.tex
\begin{table}[h]
\centering
\setlength{\tabcolsep}{10pt}
\caption{\textbf{Per-token inference latency} (ms/token) of different decoding strategies. We report the mean ± standard deviation over 300 samples.}
\vspace{0.1in}
\label{tab:computation}
\begin{small}
\begin{tabular}{l|cc}
    \toprule
    \textbf{Method} & \textbf{LLaVA-1.5-7B} & \textbf{Qwen2.5-VL-7B} \\
    \midrule
    Greedy & \tc{26.0 ± 0.2} & \tc{72.3 ± 1.5} \\
    DoLa   & 34.2 ± 0.4 & 82.6 ± 0.4 \\
    VCD    & 51.7 ± 0.4 & 143.4 ± 1.1 \\
    M3ID   & 49.5 ± 0.2 & 143.3 ± 13.2 \\
    CODE   & 81.0 ± 6.3 & 176.2 ± 45.1 \\
    SID    & 51.9 ± 0.4 & 146.5 ± 1.3 \\
    \ca{Ours}   & \ca{26.5 ± 0.2} & \ca{72.7 ± 0.4} \\
    \bottomrule
\end{tabular}
\end{small}
\end{table}

%% file: Tables/Ablation.tex
\begin{table}[h]
    \renewcommand{\arraystretch}{0.9}
    \setlength{\tabcolsep}{5pt}
    \small
    \centering
    \caption{\textbf{VQAv2 ablation results.}
    Accuracy (\%) under different choices of layer scope and threshold $\alpha$ for Qwen2.5-VL-7B and InternVL3-8B.
    Higher scores ($\uparrow$) indicate better performance.}
    \vspace{0.05in}
    \label{tab:ablation_vqa}
    \begin{tabular}{c c | c c}
        \toprule
        \textbf{Layer Scope} & $\boldsymbol{\alpha}$ & \textbf{Qwen2.5-VL-7B} & \textbf{InternVL3-8B} \\
        \midrule
        last & $10^{-1}$ & 65.73 & 67.60 \\
        last & $10^{-2}$ & 65.13 & 66.93 \\
        last & $10^{-3}$ & 65.53 & 64.53 \\
        last & $10^{-4}$ & 62.60 & 63.40 \\
        all  & $10^{-1}$ & 65.93 & 68.40 \\
        \cc{all} & \cc{$10^{-2}$} & \cc{67.60} & \cc{68.13} \\
        all  & $10^{-3}$ & 66.27 & 64.87 \\
        all  & $10^{-4}$ & 64.00 & 62.27 \\
        \bottomrule
    \end{tabular}
\end{table}

\begin{table*}[h]
    \setlength{\tabcolsep}{10pt}
    \small
    \centering
    \caption{\textbf{Ablation study.}
    Panel (a) varies the vision token selection criterion (\textit{min/max} = min-/max-JSD, \textit{random} = uniform) and use of a vocabulary subset (\textit{full} = no constraint, \textit{subset} = vocabulary subset constraint).
    Panel (b) varies the layer scope (\textit{last} vs \textit{all}) and the threshold $\alpha$.
    Panel (c) compares alternative selection metrics while fixing all other components.
    Panel (d) compares fusion design choices under the same constrained candidate set.
    Lower scores ($\downarrow$) on CHAIR$_S$, CHAIR$_I$ and higher ($\uparrow$) F1 indicate better performance.
    Highlighted rows mark the main configuration.
    }
    \label{tab:ablation}
    \begin{tabular}{c c | c c c}
        \toprule
        \multicolumn{5}{c}{\textbf{(a) Vocabulary Constraint $\times$ Token Selection}}\\
        \midrule
        \textbf{Vocab} & \textbf{Selection} & \textbf{CHAIR$_S$} $\downarrow$ & \textbf{CHAIR$_I$} $\downarrow$ & \textbf{F1} $\uparrow$ \\
        \midrule
        subset  & max     & 0.0 & 0.00 & 0.67 \\
        subset  & random  & 2.6 & 7.94 & 17.63 \\
        \cc{subset}  & \cc{min}     & \cc{29.8} & \cc{7.04} & \cc{81.16} \\
        full & min     & 0.2 & 7.14 & 1.52 \\
        \toprule
        \multicolumn{5}{c}{\textbf{(b) Layer Scope $\times$ Threshold $\alpha$}}\\
        \midrule
        \textbf{Layer Scope} & $\boldsymbol{\alpha}$ & \textbf{CHAIR$_S$} $\downarrow$ & \textbf{CHAIR$_I$} $\downarrow$ & \textbf{F1} $\uparrow$ \\
        \midrule
        last & $1\mathrm{e}{-1}$ & 33.8 & 8.32 & 80.02 \\
        last & $1\mathrm{e}{-2}$ & 37.0 & 9.13 & 79.88 \\
        last & $1\mathrm{e}{-3}$ & 34.0 & 7.88 & 80.97 \\
        last & $1\mathrm{e}{-4}$ & 34.2 & 8.21 & 80.41 \\
        last & $1\mathrm{e}{-5}$ & 31.4 & 8.04 & 80.12 \\
        last & $1\mathrm{e}{-6}$ & 32.8 & 7.21 & 79.68 \\
        all & $1\mathrm{e}{-1}$ & 33.8 & 8.82 & 80.70 \\
        all & $1\mathrm{e}{-2}$ & 36.8 & 8.91 & 79.93 \\
        all & $1\mathrm{e}{-3}$ & 38.8 & 9.36 & 79.60 \\
        all & $1\mathrm{e}{-4}$ & 32.8 & 8.17 & 80.17 \\
        \cc{all} & \cc{$1\mathrm{e}{-5}$} & \cc{29.8} & \cc{7.04} & \cc{81.16} \\
        all & $1\mathrm{e}{-6}$ & 32.4 & 9.38 & 79.90 \\
        \toprule
        \multicolumn{5}{c}{\textbf{(c) Selection Metric Comparison}}\\
        \midrule
        \textbf{Selection Metric} &  & \textbf{CHAIR$_S$} $\downarrow$ & \textbf{CHAIR$_I$} $\downarrow$ & \textbf{F1} $\uparrow$ \\
        \midrule
        KL     &  & 32.0 & 7.62 & 79.09 \\
        Cosine &  & 32.4 & 8.07 & 80.16 \\
        \cc{JSD} & \cc{ } & \cc{29.8} & \cc{7.04} & \cc{81.16} \\
        \toprule
        \multicolumn{5}{c}{\textbf{(d) Fusion Design Comparison}}\\
        \midrule
        \textbf{Fusion} & \textbf{Weight} & \textbf{CHAIR$_S$} $\downarrow$ & \textbf{CHAIR$_I$} $\downarrow$ & \textbf{F1} $\uparrow$ \\
        \midrule
        Interpolation & 0.25 & 35.6 & 9.13 & 79.72 \\
        Interpolation & 0.5  & 32.2 & 8.75 & 81.00 \\
        Interpolation & 0.75 & 31.2 & 9.68 & 80.22 \\
        Product-of-Experts           & 0.25 & 33.4 & 8.07 & 80.25 \\
        Product-of-Experts           & 0.5  & 32.0 & 7.73 & 80.35 \\
        \cc{Product-of-Experts} & \cc{1.0} & \cc{29.8} & \cc{7.04} & \cc{81.16} \\
        Product-of-Experts           & 2.0  & 33.2 & 9.11 & 79.86 \\
        \bottomrule
    \end{tabular}
    \vspace{0.5in}
\end{table*}

%% file: Tables/appendix_fig1.tex
\begin{table*}[b]
\centering
\caption{
    \textbf{Detailed numerical values for Figure~\ref{fig:overview}.} Token scores over the constrained vocabulary $\mathcal{V}_{\texttt{cons}}^2$ at decoding step $t=2$ with \name{} are presented. The selected vision token (index \texttt{229}) amplifies the relevance of ``\texttt{three}'', increasing its probability from 35.45\% (base) to 58.27\% (final), while suppressing ``\texttt{four}'' from 50.78\% to 38.21\%.
}
\vspace{0.1in}
\scriptsize
\renewcommand{\arraystretch}{1.1}
\begin{tabular}{|r|l|rr|rr|rr|}
    \hline
    \textbf{Rank} & \textbf{Token} 
    & \multicolumn{2}{c|}{\textbf{Base Logit}} 
    & \multicolumn{2}{c|}{\textbf{(\texttt{229}) Vision Token Logit}} 
    & \multicolumn{2}{c|}{\textbf{Final Logit}} \\
    \cline{3-8}
    & & \textbf{Log-Prob} & \textbf{Prob (\%)} 
      & \textbf{Log-Prob} & \textbf{Prob (\%)} 
      & \textbf{Log-Prob} & \textbf{Prob (\%)} \\
    \hline
      1 & three      & $ -1.04$ & $35.45$ & $ -1.47$ & $22.98$ & $ -2.51$ & $58.27$ \\
      2 & four       & $ -0.68$ & $50.78$ & $ -2.25$ & $10.52$ & $ -2.93$ & $38.21$ \\
      3 & two        & $ -4.57$ & $ 1.04$ & $ -1.19$ & $30.44$ & $ -5.76$ & $ 2.26$ \\
      4 & five       & $ -2.54$ & $ 7.91$ & $ -4.45$ & $ 1.16$ & $ -6.99$ & $ 0.66$ \\
      5 & a          & $ -4.33$ & $ 1.31$ & $ -3.08$ & $ 4.58$ & $ -7.42$ & $ 0.43$ \\
      6 & six        & $ -3.94$ & $ 1.94$ & $ -5.15$ & $ 0.58$ & $ -9.10$ & $ 0.08$ \\
      7 &            & $ -5.10$ & $ 0.61$ & $ -4.22$ & $ 1.47$ & $ -9.32$ & $ 0.06$ \\
      8 & several    & $ -5.75$ & $ 0.32$ & $ -5.39$ & $ 0.45$ & $-11.15$ & $ 0.01$ \\
      9 & seven      & $ -5.79$ & $ 0.31$ & $ -6.12$ & $ 0.22$ & $-11.90$ & $ 0.00$ \\
     10 & un         & $ -8.65$ & $ 0.02$ & $ -3.80$ & $ 2.24$ & $-12.45$ & $ 0.00$ \\
     11 & at         & $ -8.28$ & $ 0.03$ & $ -4.77$ & $ 0.85$ & $-13.05$ & $ 0.00$ \\
     12 & one        & $-10.05$ & $ 0.00$ & $ -3.69$ & $ 2.50$ & $-13.74$ & $ 0.00$ \\
     13 & in         & $-11.15$ & $ 0.00$ & $ -2.99$ & $ 5.01$ & $-14.14$ & $ 0.00$ \\
     14 & eight      & $ -7.38$ & $ 0.06$ & $ -6.78$ & $ 0.11$ & $-14.16$ & $ 0.00$ \\
     15 & f          & $ -9.52$ & $ 0.01$ & $ -4.76$ & $ 0.86$ & $-14.28$ & $ 0.00$ \\
     16 & multiple   & $ -7.66$ & $ 0.05$ & $ -6.66$ & $ 0.13$ & $-14.32$ & $ 0.00$ \\
     17 & only       & $ -8.68$ & $ 0.02$ & $ -5.89$ & $ 0.28$ & $-14.57$ & $ 0.00$ \\
     18 & nine       & $ -7.71$ & $ 0.04$ & $ -6.97$ & $ 0.09$ & $-14.68$ & $ 0.00$ \\
     19 & half       & $ -8.62$ & $ 0.02$ & $ -6.22$ & $ 0.20$ & $-14.85$ & $ 0.00$ \\
     20 & many       & $ -8.70$ & $ 0.02$ & $ -6.21$ & $ 0.20$ & $-14.91$ & $ 0.00$ \\
     21 & all        & $-10.38$ & $ 0.00$ & $ -4.85$ & $ 0.78$ & $-15.23$ & $ 0.00$ \\
     22 & still      & $-11.11$ & $ 0.00$ & $ -4.15$ & $ 1.57$ & $-15.26$ & $ 0.00$ \\
     23 & an         & $-10.33$ & $ 0.00$ & $ -5.10$ & $ 0.61$ & $-15.43$ & $ 0.00$ \\
     24 & some       & $-10.05$ & $ 0.00$ & $ -5.40$ & $ 0.45$ & $-15.46$ & $ 0.00$ \\
     25 & more       & $-11.15$ & $ 0.00$ & $ -4.68$ & $ 0.92$ & $-15.83$ & $ 0.00$ \\
     26 & about      & $-10.05$ & $ 0.00$ & $ -5.90$ & $ 0.27$ & $-15.95$ & $ 0.00$ \\
     27 & both       & $-10.74$ & $ 0.00$ & $ -5.23$ & $ 0.53$ & $-15.98$ & $ 0.00$ \\
     28 & the        & $-11.98$ & $ 0.00$ & $ -4.07$ & $ 1.71$ & $-16.05$ & $ 0.00$ \\
     29 & not        & $-11.58$ & $ 0.00$ & $ -4.59$ & $ 1.02$ & $-16.16$ & $ 0.00$ \\
     30 & over       & $-11.99$ & $ 0.00$ & $ -4.25$ & $ 1.42$ & $-16.25$ & $ 0.00$ \\
     31 & different  & $-11.93$ & $ 0.00$ & $ -4.73$ & $ 0.88$ & $-16.66$ & $ 0.00$ \\
     32 & ten        & $ -8.56$ & $ 0.02$ & $ -8.15$ & $ 0.03$ & $-16.71$ & $ 0.00$ \\
     33 & just       & $-10.97$ & $ 0.00$ & $ -5.78$ & $ 0.31$ & $-16.75$ & $ 0.00$ \\
     34 & fruit      & $-11.81$ & $ 0.00$ & $ -5.04$ & $ 0.65$ & $-16.85$ & $ 0.00$ \\
     35 & as         & $-12.06$ & $ 0.00$ & $ -4.82$ & $ 0.81$ & $-16.88$ & $ 0.00$ \\
     36 & small      & $-12.11$ & $ 0.00$ & $ -4.86$ & $ 0.78$ & $-16.97$ & $ 0.00$ \\
     37 & various    & $-10.39$ & $ 0.00$ & $ -6.71$ & $ 0.12$ & $-17.10$ & $ 0.00$ \\
     38 & also       & $-11.94$ & $ 0.00$ & $ -5.21$ & $ 0.55$ & $-17.15$ & $ 0.00$ \\
     39 & very       & $-11.80$ & $ 0.00$ & $ -5.54$ & $ 0.39$ & $-17.35$ & $ 0.00$ \\
     40 & around     & $-11.28$ & $ 0.00$ & $ -6.07$ & $ 0.23$ & $-17.35$ & $ 0.00$ \\
     41 & numerous   & $-10.02$ & $ 0.00$ & $ -7.51$ & $ 0.05$ & $-17.54$ & $ 0.00$ \\
     42 & no         & $-10.70$ & $ 0.00$ & $ -6.93$ & $ 0.10$ & $-17.64$ & $ 0.00$ \\
     43 & total      & $-11.17$ & $ 0.00$ & $ -6.80$ & $ 0.11$ & $-17.97$ & $ 0.00$ \\
     44 & currently  & $-11.17$ & $ 0.00$ & $ -6.89$ & $ 0.10$ & $-18.06$ & $ 0.00$ \\
     45 & cut        & $-11.78$ & $ 0.00$ & $ -6.46$ & $ 0.16$ & $-18.24$ & $ 0.00$ \\
     46 & quite      & $-11.62$ & $ 0.00$ & $ -6.71$ & $ 0.12$ & $-18.33$ & $ 0.00$ \\
     47 & few        & $-11.91$ & $ 0.00$ & $ -6.64$ & $ 0.13$ & $-18.55$ & $ 0.00$ \\
     48 & twelve     & $-10.07$ & $ 0.00$ & $ -8.53$ & $ 0.02$ & $-18.60$ & $ 0.00$ \\
     49 & lots       & $-11.52$ & $ 0.00$ & $ -7.37$ & $ 0.06$ & $-18.89$ & $ 0.00$ \\
     50 & th         & $-11.34$ & $ 0.00$ & $ -7.63$ & $ 0.05$ & $-18.98$ & $ 0.00$ \\
     51 & actually   & $-11.84$ & $ 0.00$ & $ -7.40$ & $ 0.06$ & $-19.24$ & $ 0.00$ \\
     52 & approximately & $-11.05$ & $ 0.00$ & $ -8.45$ & $ 0.02$ & $-19.50$ & $ 0.00$ \\
     53 & exactly    & $-12.14$ & $ 0.00$ & $ -7.48$ & $ 0.06$ & $-19.62$ & $ 0.00$ \\
     54 & plenty     & $-11.78$ & $ 0.00$ & $ -7.95$ & $ 0.04$ & $-19.73$ & $ 0.00$ \\
     55 & eleven     & $-10.82$ & $ 0.00$ & $ -9.25$ & $ 0.01$ & $-20.07$ & $ 0.00$ \\
    \hline
\end{tabular}
\vspace{0.2cm}
\label{tab:appendix_fig1}
\end{table*}

%% file: Tables/appendix_fig2_1.tex
\begin{table*}[b]
\centering
\caption{
    \textbf{Detailed numerical values for Figure~\ref{fig:motivation} (left).} Log-probabilities and probabilities (\%) of selected constrained vocabulary tokens as projected by three vision tokens (pink, yellow, cyan) under the full vocabulary $\mathcal{V}$ and the constrained vocabulary $\mathcal{V}_{\texttt{env}}$.
}
\vspace{0.1in}
\scriptsize
\renewcommand{\arraystretch}{1.2}
\resizebox{\linewidth}{!}{
\begin{tabular}{|l|
                    rr|rr|
                    rr|rr|
                    rr|rr|}
    \hline
    \textbf{Vision Token}
    & \multicolumn{4}{c|}{\textbf{Top (Pink)}} 
    & \multicolumn{4}{c|}{\textbf{Middle (Yellow)}} 
    & \multicolumn{4}{c|}{\textbf{Bottom (Cyan)}} \\
    \hline
    \textbf{Vocab}
    & \multicolumn{2}{c|}{$\mathcal{V}$} & \multicolumn{2}{c|}{$\mathcal{V}_{\texttt{env}}$} 
    & \multicolumn{2}{c|}{$\mathcal{V}$} & \multicolumn{2}{c|}{$\mathcal{V}_{\texttt{env}}$} 
    & \multicolumn{2}{c|}{$\mathcal{V}$} & \multicolumn{2}{c|}{$\mathcal{V}_{\texttt{env}}$}  \\
    \hline
    \textbf{Token}
    & \textbf{Log-Prob} & \textbf{\%} 
    & \textbf{Log-Prob} & \textbf{\%} 
    & \textbf{Log-Prob} & \textbf{\%} 
    & \textbf{Log-Prob} & \textbf{\%} 
    & \textbf{Log-Prob} & \textbf{\%} 
    & \textbf{Log-Prob} & \textbf{\%}  \\
    \hline
    river      & $-14.74$ & $  0.00$ & $ -6.16$ & $  0.21$ & $ -3.61$ & $  2.70$ & \ca{$ -0.36$} & \ca{$ 69.75$} & $-14.58$ & $  0.00$ & $ -5.02$ & $  0.66$ \\
    mountain   & $-14.07$ & $  0.00$ & $ -5.48$ & $  0.42$ & $ -6.47$ & $  0.15$ & $ -3.22$ & $  3.99$ & $-15.02$ & $  0.00$ & $ -5.46$ & $  0.43$ \\
    person     & $-13.39$ & $  0.00$ & $ -4.80$ & $  0.82$ & $-12.27$ & $  0.00$ & $ -9.01$ & $  0.01$ & $ -9.86$ & $  0.01$ & \ca{$ -0.30$} & \ca{$ 73.81$} \\
    sky        & $ -8.66$ & $  0.02$ & \ca{$ -0.07$} & \ca{$ 93.44$} & $ -6.76$ & $  0.12$ & $ -3.51$ & $  2.99$ & $-12.82$ & $  0.00$ & $ -3.26$ & $  3.83$ \\
    tree       & $-12.06$ & $  0.00$ & $ -3.48$ & $  3.09$ & $-11.71$ & $  0.00$ & $ -8.46$ & $  0.02$ & $-12.47$ & $  0.00$ & $ -2.92$ & $  5.42$ \\
    cloud      & $-13.88$ & $  0.00$ & $ -5.30$ & $  0.50$ & $ -9.52$ & $  0.01$ & $ -6.26$ & $  0.19$ & $-12.46$ & $  0.00$ & $ -2.91$ & $  5.47$ \\
    sea        & $-13.80$ & $  0.00$ & $ -5.21$ & $  0.55$ & $ -6.82$ & $  0.11$ & $ -3.57$ & $  2.81$ & $-13.24$ & $  0.00$ & $ -3.68$ & $  2.52$ \\
    grass      & $-15.41$ & $  0.00$ & $ -6.82$ & $  0.11$ & $ -7.19$ & $  0.08$ & $ -3.94$ & $  1.95$ & $-13.41$ & $  0.00$ & $ -3.86$ & $  2.11$ \\
    rock       & $-13.35$ & $  0.00$ & $ -4.77$ & $  0.85$ & $ -4.95$ & $  0.71$ & $ -1.70$ & $ 18.33$ & $-12.41$ & $  0.00$ & $ -2.85$ & $  5.77$ \\
    \hline
\end{tabular}
}
\vspace{0.2cm}
\label{tab:appendix_fig2_1}
\end{table*}

%% file: Tables/appendix_fig2_2.tex
\begin{table*}[b]
\centering
\caption{
    \textbf{Detailed numerical values for Figure~\ref{fig:motivation} (right).} Token scores over the constrained vocabulary $\mathcal{V}_{\texttt{cons}}^{72}$ at decoding step $t=72$ with \name{}. The selected vision token (index \texttt{$193$}) amplifies the relevance of ``\texttt{painting}'', increasing its probability from 8.08\% (base) to 39.15\% (final) while suppressing hallucinatory token ``\texttt{person}'' from 15.10\% to 7.19\%.
}
\vspace{0.1in}
\scriptsize
\renewcommand{\arraystretch}{1.1}
\begin{tabular}{|r|l|rr|rr|rr|}
    \hline
    \textbf{Rank} & \textbf{Token} 
    & \multicolumn{2}{c|}{\textbf{Base Logit}} 
    & \multicolumn{2}{c|}{\textbf{(\texttt{193}) Vision Token Logit}} 
    & \multicolumn{2}{c|}{\textbf{Final Logit}} \\
    \cline{3-8}
    & & \textbf{Log-Prob} & \textbf{Prob (\%)} 
      & \textbf{Log-Prob} & \textbf{Prob (\%)} 
      & \textbf{Log-Prob} & \textbf{Prob (\%)} \\
    \hline
      1 & painting   & $ -2.52$ & $ 8.08$ & $ -1.27$ & $27.99$ & $ -3.79$ & $39.15$ \\
      2 & rock       & $ -2.60$ & $ 7.42$ & $ -2.28$ & $10.22$ & $ -4.88$ & $13.11$ \\
      3 & mountain   & $ -2.14$ & $11.76$ & $ -3.03$ & $ 4.83$ & $ -5.17$ & $ 9.82$ \\
      4 & person     & $ -1.89$ & $15.10$ & $ -3.59$ & $ 2.75$ & $ -5.48$ & $ 7.19$ \\
      5 & landscape  & $ -2.27$ & $10.30$ & $ -3.42$ & $ 3.27$ & $ -5.70$ & $ 5.82$ \\
      6 & d          & $ -3.45$ & $ 3.16$ & $ -2.29$ & $10.14$ & $ -5.74$ & $ 5.55$ \\
      7 & p          & $ -3.91$ & $ 2.01$ & $ -2.45$ & $ 8.60$ & $ -6.36$ & $ 3.00$ \\
      8 & scene      & $ -3.80$ & $ 2.23$ & $ -2.70$ & $ 6.75$ & $ -6.50$ & $ 2.60$ \\
      9 & small      & $ -2.83$ & $ 5.91$ & $ -3.70$ & $ 2.48$ & $ -6.52$ & $ 2.54$ \\
     10 & beautiful  & $ -3.67$ & $ 2.54$ & $ -3.03$ & $ 4.84$ & $ -6.70$ & $ 2.13$ \\
     11 & large      & $ -2.93$ & $ 5.34$ & $ -3.78$ & $ 2.28$ & $ -6.71$ & $ 2.11$ \\
     12 & scen       & $ -3.57$ & $ 2.81$ & $ -3.17$ & $ 4.19$ & $ -6.74$ & $ 2.04$ \\
     13 & hill       & $ -4.01$ & $ 1.82$ & $ -2.83$ & $ 5.91$ & $ -6.84$ & $ 1.86$ \\
     14 & chair      & $ -2.85$ & $ 5.78$ & $ -4.38$ & $ 1.26$ & $ -7.23$ & $ 1.26$ \\
     15 & bow        & $ -3.87$ & $ 2.09$ & $ -3.68$ & $ 2.53$ & $ -7.54$ & $ 0.92$ \\
     16 & second     & $ -3.80$ & $ 2.23$ & $ -4.52$ & $ 1.09$ & $ -8.33$ & $ 0.42$ \\
     17 & smaller    & $ -3.02$ & $ 4.90$ & $ -6.04$ & $ 0.24$ & $ -9.06$ & $ 0.20$ \\
     18 & boat       & $ -3.00$ & $ 4.98$ & $ -6.34$ & $ 0.18$ & $ -9.34$ & $ 0.15$ \\
     19 & distant    & $ -4.14$ & $ 1.59$ & $ -5.43$ & $ 0.44$ & $ -9.57$ & $ 0.12$ \\
    \hline
\end{tabular}
\vspace{0.2cm}
\label{tab:appendix_fig2_2}
\end{table*}

%% file: Figures/quali_noAPC.tex
\begin{figure*}[b]
\centering
\includegraphics[width=1.0\textwidth]{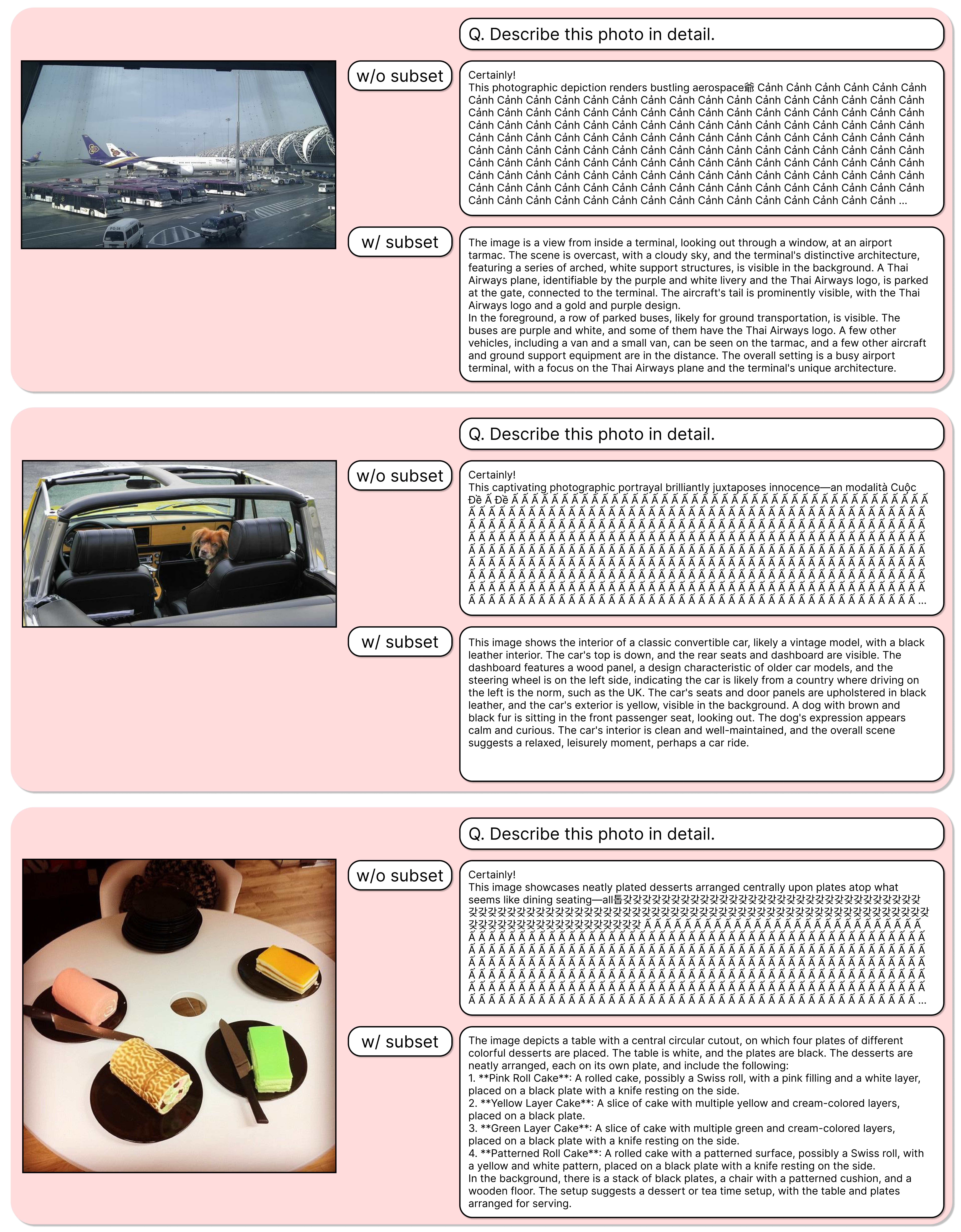}
\caption{
    \textbf{Without vocabulary subset case study.} 
    Qualitative case study from Qwen2.5-VL-7B. 
    \texttt{w/o subset} refers to ablation result of \textit{without vocabulary subset constraint}, whereas \texttt{w/ subset} refers to our proposed \name{}.
}
\label{fig:noAPC}
\end{figure*}

%% file: Figures/quali_ex1.tex
\begin{figure*}[b]
\centering
\includegraphics[width=1.0\textwidth]{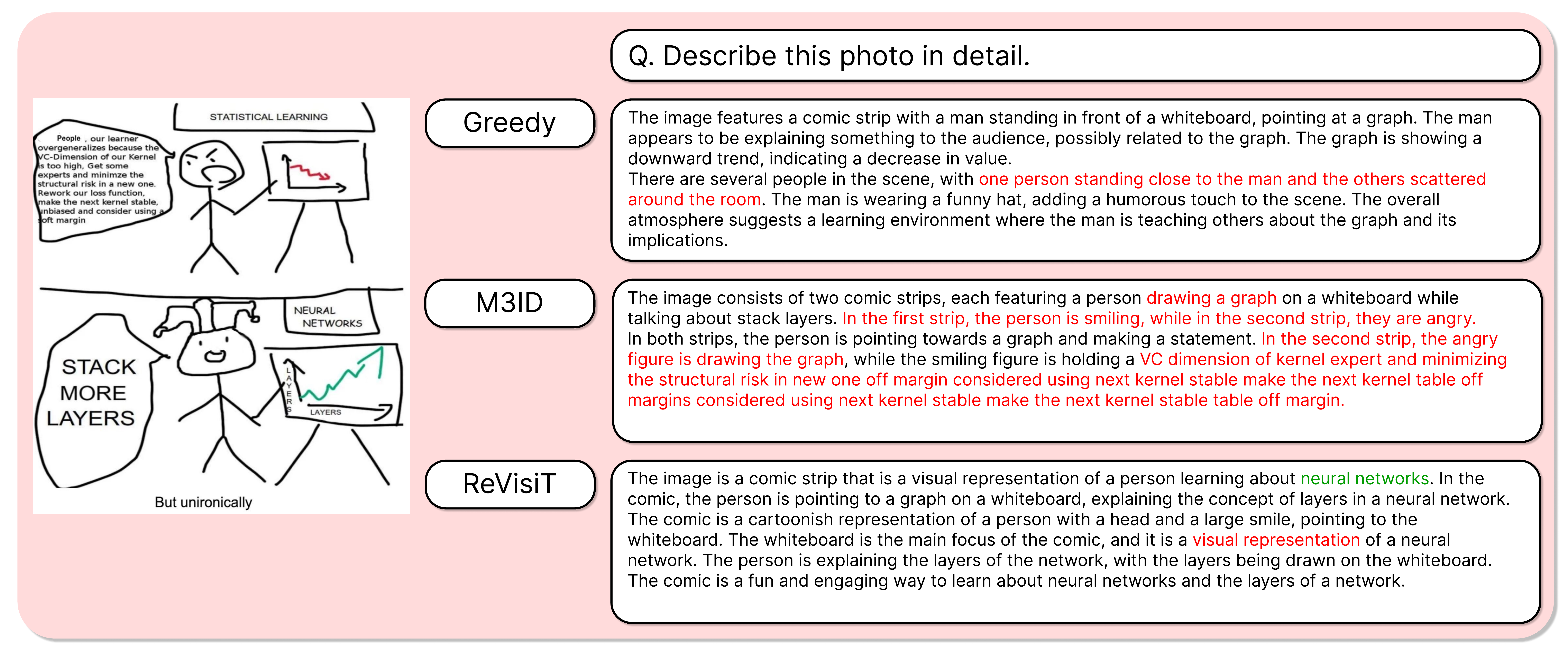}
\caption{
    \textbf{Qualitative example.} 
    The input image is a cartoon-style illustration contrasting classical statistical learning and neural network reasoning via a visual metaphor, emphasizing the shift from theoretical rigor to the heuristic of ``stacking more layers'' 
    We compare the generated responses of vanilla greedy decoding, M3ID, and \name{}, highlighting how \name{} better captures the intended visual analogy compared to the baseline methods.
}
\label{fig:quali_ex1}
\end{figure*}

%% file: Figures/quali_ex2.tex
\begin{figure*}[b]
\centering
\includegraphics[width=1.0\textwidth]{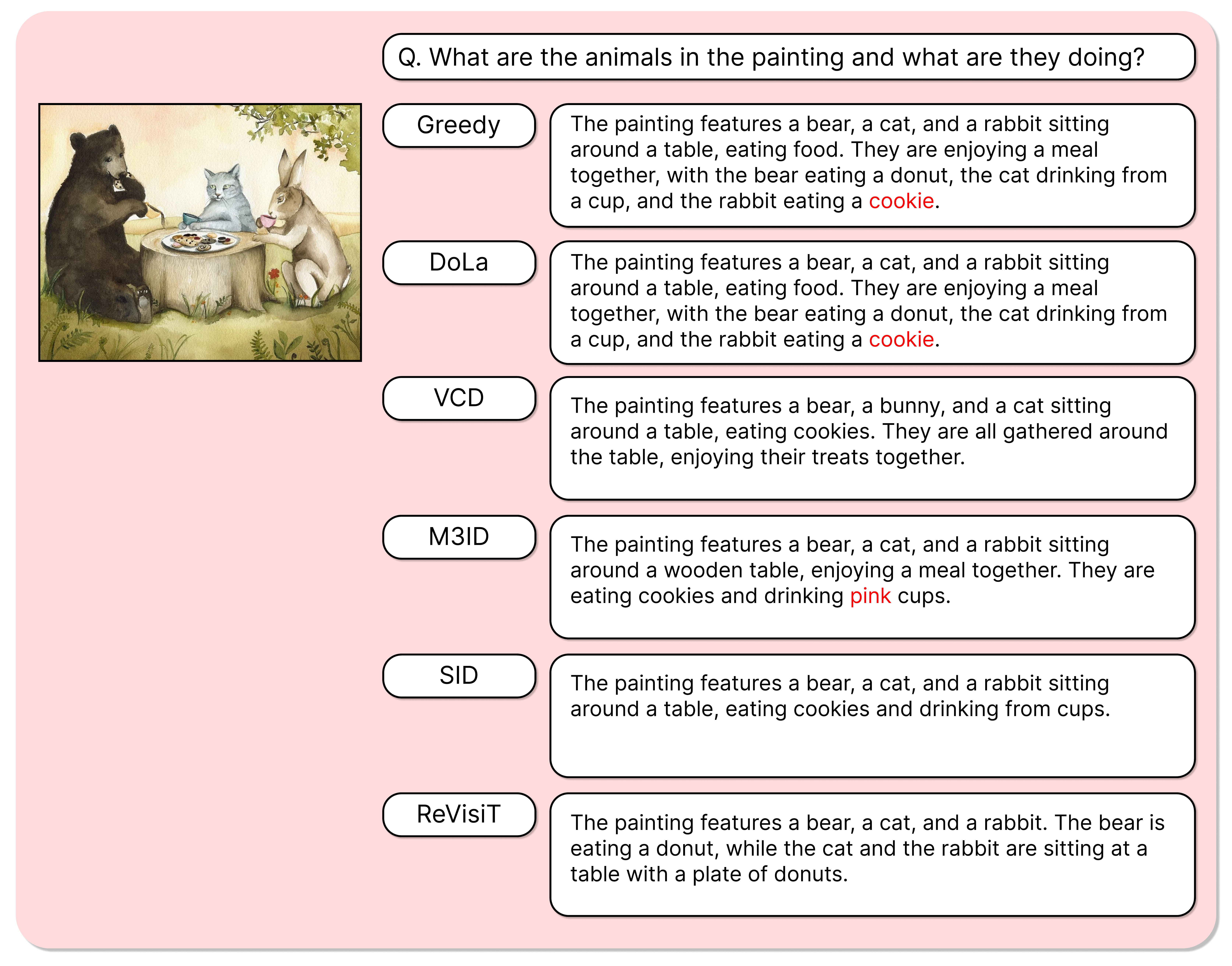}
\caption{
    \textbf{Qualitative comparison with baselines.} 
    The input image is a illustration showing a bear, a cat, and a rabbit seated around a table with a plate of donuts. 
    We compare the responses of vanilla greedy decoding and \name{} to the question, \textit{``What are the animals in the painting and what are they doing?''} 
    While the greedy output introduces a hallucinated detail (``\texttt{cookie}'') and assigns actions not visually supported (\textit{e.g., ``the cat drinking from a cup''}), \name{} provides a more faithful description aligned with the visual content, accurately identifying the animals and their activities.
}
\label{fig:quali_ex2}
\end{figure*}